%% file: main.tex
\documentclass{article}

\usepackage{PRIMEarxiv}

\usepackage[utf8]{inputenc} 
\usepackage[T1]{fontenc}    
\usepackage{hyperref}       
\usepackage{url}            
\usepackage{booktabs}       
\usepackage{amsfonts}       
\usepackage{nicefrac}       
\usepackage{microtype}      
\usepackage{fancyhdr}       
\usepackage{graphicx}       
\usepackage{caption}        
\usepackage{svg}
\usepackage{float}
\usepackage[table]{xcolor}
\usepackage{pdf14}
\graphicspath{{media/}{figures/}}     

\input{math_commands.tex}

\input{macros}

\title{\textit{ShapeGaussian}: High-Fidelity 4D Human Reconstruction in Monocular Videos via Vision Priors
}

\author{
  \textbf{Zhenxiao Liang}$^{1}$, \textbf{Ning Zhang}$^{2}$, \textbf{Youbao Tang}$^{2}$, \textbf{Ruei-Sung Lin}$^{2}$, \\
  \textbf{Qixing Huang}$^{1}$, \textbf{Peng Chang}$^{2}$, \textbf{Jing Xiao}$^{2}$ \\[1em]
  $^{1}$The University of Texas at Austin \\
  $^{2}$PAII Inc.
}

\begin{document}
\maketitle

\begin{abstract}
We introduce \textit{ShapeGaussian}, a high-fidelity, template-free method for 4D human reconstruction from casual monocular videos. Generic reconstruction methods lacking robust vision priors, such as 4DGS, struggle to capture high-deformation human motion without multi-view cues. While template-based approaches, primarily relying on SMPL, such as HUGS, can produce photorealistic results, they are highly susceptible to errors in human pose estimation, often leading to unrealistic artifacts. In contrast, ShapeGaussian effectively integrates template-free vision priors to achieve both high-fidelity and robust scene reconstructions. Our method follows a two-step pipeline: first, we learn a coarse, deformable geometry using pretrained models that estimate data-driven priors, providing a foundation for reconstruction. Then, we refine this geometry using a neural deformation model to capture fine-grained dynamic details. By leveraging 2D vision priors, we mitigate artifacts from erroneous pose estimation in template-based methods and employ multiple reference frames to resolve the invisibility issue of 2D keypoints in a template-free manner. Extensive experiments demonstrate that ShapeGaussian surpasses template-based methods in reconstruction accuracy, achieving superior visual quality and robustness across diverse human motions in casual monocular videos.
\end{abstract}

\begin{center}
    \centering
    \captionsetup{type=figure}
    \includegraphics[width=\textwidth]{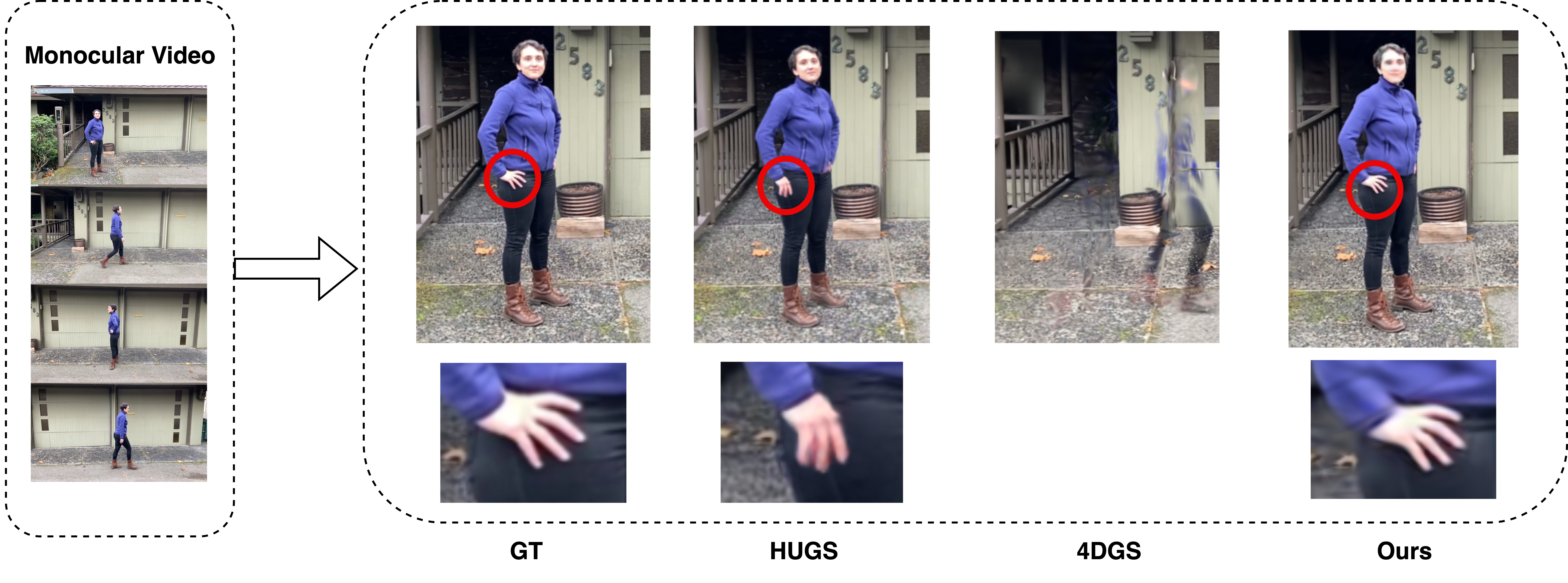}
    \captionof{figure}{We propose \textit{ShapeGaussian} to achieve high-fidelity dynamic scene reconstruction from monocular, human-centric videos in a template-free manner. Generic reconstruction methods without robust vision priors, such as 4DGS~\cite{wu_4d_2023}, struggle to capture high-deformation human motion without multi-view cues. Although template-based approaches (primarily SMPL~\cite{loper_smpl_2015}) such as HUGS~\cite{kocabas_hugs_2023} can produce photorealistic results, they are susceptible to errors in human pose estimation, often leading to unrealistic artifacts that compromise applicability. In contrast, our method delivers both high-fidelity and high-quality scene reconstructions by incorporating template-free vision priors effectively.}
\label{fig:teaser}
\end{center}

\input{sec/sec-introduction}
\input{sec/sec-related}
\input{sec/sec-preliminaries}
\input{sec/sec-method}
\input{sec/sec-experiments}

\input{sec/sec-conclusion}

\bibliographystyle{unsrt}
\bibliography{references}

\newpage
\appendix
\input{sec/sec-appendix}

\end{document}

%% file: math_commands.tex
\usepackage{amsmath,amsfonts,bm}

\def\eqref#1{equation~\ref{#1}}

\def\1{\bm{1}}

\DeclareMathAlphabet{\mathsfit}{\encodingdefault}{\sfdefault}{m}{sl}
\SetMathAlphabet{\mathsfit}{bold}{\encodingdefault}{\sfdefault}{bx}{n}

\newcommand{\R}{\mathbb{R}}

\newcommand{\Var}{\mathrm{Var}}



%% file: macros.tex
\let \bs=\boldsymbol

\def \saliency {\textup{\saliency}}

\def \path {\mathit{path}}

\date{}

 \global\long\def\R{\mathbb{R}}

%% file: sec/sec-introduction.tex
\section{Introduction}

Reconstruction of photo-realistic dynamic 3D scenes is an active research topic in computer vision, with broad applications ranging from video game synthesis to robotic manipulation. Recently, significant progress has been accomplished to recover this 4D representation from synchronized multi-view videos using Gaussian Splatting~\cite{kerbl_3d_2023,yang_deformable_2023,bae_per-gaussian_2024,lin_gaussian-flow_2023,luiten_dynamic_2023}. Yet, such specific set-up limits its applicability for daily, causal usage. In contrast, 4D reconstruction from a monocular video has more pratical appeal.

A significant limitation of existing Gaussian-based reconstruction methods is their poor performance on monocular video captures featuring rapid human motion, particularly in novel view synthesis. This deficiency primarily stems from the inaccurately reconstructed geometry of the Gaussians, compounded by the unconstrained nature of the optimization problem due to the limited multi-view cues available from monocular captures. Although recent efforts have attempted to address this issue by incorporating explicit mesh-based priors into monocular video-based Gaussian avatars, these methods still struggle with reconstructing loose clothing and hair, as well as the variable and smooth appearance of the human subject over time.



State-of-the-art approaches~\cite{huang_one-shot_2023,svitov_haha_2024,lei_gart_2023,kocabas_hugs_2023,jung_deformable_2023,li_gaussianbody_2024} for 4D human Gaussian Splatting are based on SMPL~\cite{loper_smpl_2015}, a parametric 3D model of the human body. With this strong prior knowledge of 3D human shape, the result of 4D Gaussian Splatting from a monocular video is greatly improved. However, SMPL has its limitations. It doesn’t fit some special local shape such as hair contour, loose clothing etc. In addition, SMPL describes statistical human shapes, so it won’t fit the bodies in the video perfectly. Further, the performance of 3D human pose estimation based on SMPL deteriorates when the human subjects are strongly, partially occluded. Inaccurate pose estimation produces inaccurate 4D Gaussian splatting with statistically estimated information, which can be unsatisfactory in reconstruction tasks requiring high fidelity.


To address these challenges, we propose ShapeGaussian, a template-free method for high-fidelity, photorealistic dynamic reconstruction of human-centric scenes. Instead of relying on templates, we leverage vision foundation models~\cite{khirodkar_sapiens_2024,yang_depth_2024} for accurate human segmentation and detailed relative depth estimation, providing rich 2.5D information per frame. To track human shapes across frames, we adopt image correspondence models like DensePose~\cite{guler_densepose_2018} for tracking dense keypoints. By incorporating dense temporal correspondences and refining 3D surface estimates, our method achieves superior reconstruction quality. During training deformation network, we use multiple reference frames to robustly track 2D keypoints, significantly improving deformation accuracy.

Specifically, we introduce a two-step method that initially learns a coarse, deformable geometry using pretrained models that estimate data-driven priors, and then utilizes this geometry as a basis for reconstruction. In the second step, we utilize a neural deformation model to capture the dynamic deformation details, building upon the dynamic coarse point template. In short, we propose \textit{ShapeGaussian}, and our contributions are as follows.
\begin{itemize}
\item We leverage 2D vision priors to address human-centric scene reconstruction from monocular videos, mitigating artifacts caused by erroneous pose estimation in explicit human template-based methods.
\item We employ multiple reference frames to resolve the invisibility issue of 2D keypoints in a template-free manner.
\item Extensive experiments demonstrate that ShapeGaussian outperformed template-based methods in the reconstruction task.
\end{itemize}

%% file: sec/sec-related.tex
\section{Related Work}
\label{sec:related work}
Given our primary objective of dynamic 3D scene reconstruction involving high-deformation human motion from monocular videos, the related work is organized into two main areas. First, we examine popular representations used to model generic dynamic scene, particularly those tailored to handle complex and deforming objects. This will be discussed in Section~\ref{subsec:dynamic gaussian}. Second, due to absence of multi-view clues in monocular videos, it is critical to obtain a superior model initialization and strong regularization from vision models. We explore works for obtaining accurate correspondences and tracking information in Section~\ref{subsec:correspondences and tracking}. 

\subsection{Dynamic Reconstruction with 3D Gaussians}
\label{subsec:dynamic gaussian}
The recent rise of 3D Gaussian Splatting
(3DGS)~\cite{kerbl_3d_2023} highlights its capability for real-time rendering, thanks to its explicit point cloud format. Recent advancements have extended 3DGS to dynamic 3D scene modeling.  Generally speaking, we can classify the representations of general dynamic scene with Gaussian Splatting into two categories.

The first approach involves using a neural network to predict deformed 3D Gaussians \cite{wu_4d_2023, yang_deformable_2023, cao_hexplane_2023, li_spacetime_2023, wang_fourier_2022, liu2024modgsdynamicgaussiansplatting}. For instance, Yang et al.\cite{yang_deformable_2023} proposed a method that predicts Gaussian property offsets from a single reference frame using an MLP network. Wu et al.\cite{wu_4d_2023} introduce a deformation field employing the efficient Hexplane representation~\cite{cao_hexplane_2023}. Liu et al.\cite{liu2024modgsdynamicgaussiansplatting} leverages an invertible MLP architecture, based on\cite{wang_tracking_2023}, to mitigate issues of missing Gaussians in single-frame references.

The second approach models Gaussian trajectories using a low-rank representation \cite{wang_shape_2024, lei2024moscadynamicgaussianfusion, yang_deformable_2023}, which effectively captures smooth deformations as combinations of basis functions. This yields strong results for casual monocular videos with largely rigid motions, given the assumption that the deformation can be decomposed efficiently.

Our method aligns with the first approach, prioritizing flexibility to accommodate high-deformation human motion. Unlike prior methods, we incorporate multiple reference frames rather than relying solely on a single frame, enhancing performance for complex human movements. Additionally, our approach preserves an explicit canonical point cloud, which supports better geometric regularization than the virtual canonical space utilized by \cite{liu2024modgsdynamicgaussiansplatting}.

\subsection{Correspondences and Tracking for Human Motion}
\label{subsec:correspondences and tracking}
Monocular 3D long-range tracking remains relatively underexplored in the literature. The most common strategy to achieve high-quality long-range 3D tracking is by lifting 2D tracking information into 3D space \cite{wang_shape_2024,lei2024moscadynamicgaussianfusion,luiten_dynamic_2023,guo_motion-aware_2024,tretschk_non-rigid_2021,zhao_particlesfm_2022}. Traditional 2D tracking approaches primarily depend on optical flow to determine point correspondences across frames~\cite{weinzaepfel_deepflow_2013,black_framework_1993,sundaram_dense_2010,brox_high_2004,fischer_flownet_2015,horn_determining_1981,lucas_iterative_1981,shi_videoflow_2023, sun_pwc-net_2018,teed_raft_2020,xu_gmflow_2022,jiang2021learningopticalflowmatches,jiang2021learningestimatehiddenmotions}, which has shown efficacy for short-term tracking. However, maintaining accurate tracking over extended video sequences remains a challenge due to cumulative error and the inherent limitations of optical flow in capturing long-range dependencies. Early methods for long-term 2D trajectory estimation often relied on hand-crafted priors to infer motion trajectories, allowing for limited temporal coherence.

Recently, renewed interest in long-term 2D tracking has led to promising results on challenging, in-the-wild video data. Some approaches rely on test-time optimization, where models refine noisy, short-range motion estimates into a consolidated global representation, producing stable long-term correspondences~\cite{neoral_mft_2023,wang_tracking_2023}. Others employ data-driven techniques~\cite{doersch_tapir_2023,harley_particle_2022,karaev_cotracker_2024}, leveraging synthetic data to train neural networks to predict long-term correspondences directly. While these methods achieve impressive results, they often struggle with high-deformation human motions and scenarios with rapid camera movement, as they are optimized for smoother, predictable motion. Moreover, the latest advanced preprocessing methods~\cite{doersch2024bootstapbootstrappedtrainingtrackinganypoint,doersch_tapir_2023,wang_tracking_2023} for capturing 2D keypoint correspondences are typically slow, often requiring more processing time than the subsequent 3D reconstruction step.

Another common strategy involves tracking optimization techniques that rely on test-time adjustment with predefined human template, typically focusing on single human and novel pose synthesis\cite{kocabas_hugs_2023,cao_openpose_2019,jiang_instantavatar_2022,jiang_instantavatar_2022-1,jung_deformable_2023,svitov_haha_2024,liu_gva_2024,shao_splattingavatar_2024,qian_3dgs-avatar_2023,hu_gaussianavatar_2023,li_animatable_2023,kocabas_hugs_2023}. Although effective for isolated, consistent human, this approach is limited in dynamic scenes involving multiple, interacting humans due to its reliance on object-specific templates and assumptions of consistent appearance.

In contrast, our method leverages a pre-trained human correspondence model to extract robust 2D tracking points from UV maps, which we then lift into 3D by aligning them with estimated depth maps. Unlike prior approaches, our method operates without template priors, making it more adaptable to complex scenes with multiple fast-moving human subjects. This enables robust and efficient 3D tracking of human motion from monocular video while maintaining flexibility across diverse and dynamic settings.

%% file: sec/sec-preliminaries.tex
\section{Preliminaries}
\label{sec:preliminaries}

In this sections, we briefly review the representation of 3D-GS~\cite{kerbl_3d_2023} in Sec.~\ref{subsec:3d-gs} and the formulation of priors used by our method in Sec.~\ref{subsec:prior}.

\subsection{3D Gaussian Splatting}
\label{subsec:3d-gs}
3D Gaussians~\cite{kerbl_3d_2023} serve as an explicit 3D scene representation through point clouds. Each Gaussian is defined by a 5-tuple $(\bs{\mu},\bs{\Sigma},\bs{s},o,\bs{c})$, where $\bs{\mu}\in\mathbb{R}^3,\bs{\Sigma}\in\mathbb{SO}(3)$ are the 3D mean and orientation  and $\bs{s}\in\mathbb{R}^3$ the scale, $o\in\R$ the opacity, and $\bs{c}\in\R^3$ the color. The rendering process would first project 3D Gaussians onto the 2D image plane. More specifically, given the world-to-camera extrinsics $\mathbf{E}$ and intrinsics $\mathbf{K}$, the projection of the 3D Gaussians can be obtained by formula
\vspace{-8pt}
\begin{align}
    \bs{\mu}'(\mathbf{K},\mathbf{E}) & :=\Pi(\mathbf{KE}\bs{\mu})\in \R^2,\nonumber \\
    \Sigma'(\mathbf{K},\mathbf{E}) &:=\mathbf{J}_{\mathbf{KE}}\Sigma_0\mathbf{J}_{\mathbf{KE}}^T\in\R^{2\times 2},
\end{align}

\vspace{-8pt}
where $\Pi$ is the perspective projection operator, and $\mathbf{J}_{\mathbf{KE}}$ is the Jacobian matrix from the affine approximation of the projective transformation determined by $\mathbf{E}$ and $\mathbf{K}$ at location $\bs{\mu}$. The projected 2D Gaussians can then be efficiently rasterized into RGB image  along with the depth map via volume rendering as
\begin{equation}
    \hat{\mathbf{I}}(\bs{p}):= \sum_{i\in H(\bs{p})} T_i\alpha_i\bs{c}_i,\quad \hat{\mathbf{D}}(\bs{p}):= \sum_{i\in H(\bs{p})} T_i\alpha_i\bs{d}_i,
\end{equation}
where $H(\bs{p})$ denotes the index set of Gaussians that intersect the ray shoot from pixel $\bs{p}$, and the equivalent opacity and transmittance is calculated by
\begin{align}
    \alpha_i := &o_i\cdot\exp\big(-\tfrac{1}{2}(\bs{p}-\bs{\mu}')^T\mathbf{\Sigma}'(\bs{p}-\bs{\mu}')\big),\nonumber \\ 
    T_i:= &\prod_{j<i}(1-\alpha_j).
\end{align}

\subsection{Data-driven Priors}
\label{subsec:prior}
We consider three types of data-driven priors produced by off-the-shelf pretrained models, which assist our model in inferring accurate geometry from videos that lack sufficient multi-view cues. For each training image $\mathbf{I}$, we have the following:

\noindent\textbf{Semantic Mask.} Represented by a map $\mathbf{M}_{\mathbf{I}}$, where $\mathbf{M}_{\mathbf{I}}(\bs{p})=1$ if and only if the pixel $\bs{p}$ is within the human silhouette.

\noindent\textbf{Depth Map.} Represented by a map $\mathbf{D}_{\mathbf{I}}$, where $\mathbf{D}_{\mathbf{I}}(\bs{p})$ indicates the distance of the point on the front-most surface from the viewpoint. In our method, two different depth maps will be obtained to cover the global and human depth maps respectively.

\noindent\textbf{Human UV Maps.} Denoted by $\bs{z}_{\mathbf{I}}(\bs{p})=(i,u,v)\in\mathbb{R}^{3}$ for pixel $\bs{p}$, where $i$ denotes the part id and $(u,v)$ is the UV coordinates associated with pixel $\bs{p}$. $i=0$ indicates the pixel is not associated with any part. 

We will explain with more details in Sec.~\ref{subsec:init} to demonstrate how to initialize the Gaussians and deformation field based on the off-the-shelf foundation vision clues above.

%% file: sec/sec-method.tex
\section{Method}

\begin{figure*}[!ht]
\centering
\includegraphics[width=\textwidth]{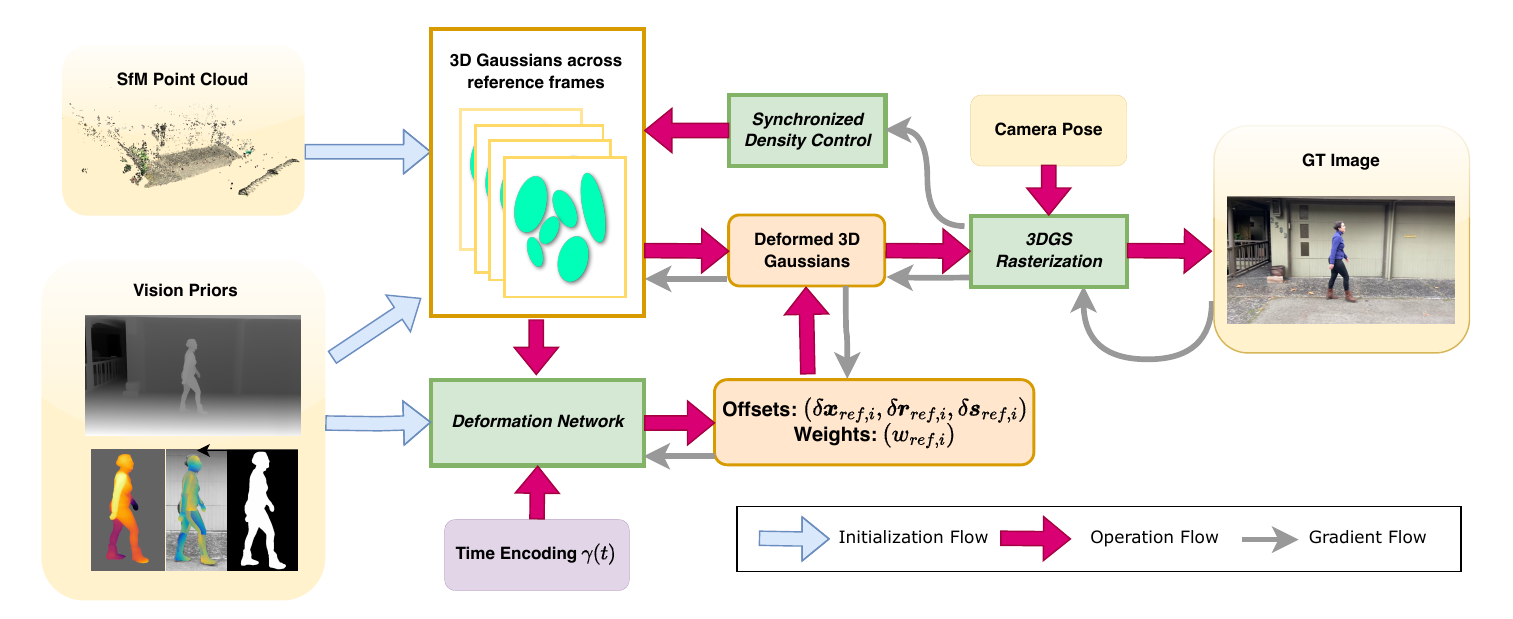}


\caption{\textbf{Overview of our proposed method.} Initialization of the deformation network and 3D Gaussians in the reference frames is performed as described in Sec.~\ref{subsec:init}. Following this, a joint optimization is conducted to refine both the deformation network and 3D Gaussians with synchronized density control across all frames.\\
Yellow boxes denote all input elements, orange boxes indicate intermediate results, and green boxes represent the actions being taken.
}
\label{fig:method}
\end{figure*}

Our method is outlined in Fig.~\ref{fig:method}. We begin by constructing a coarse dynamic Gaussian model based on vision priors derived from the input monocular video. An initial deformation field is then established to encode geometry priors. Following this, we optimize the deformation field jointly with Gaussians among reference frames, incorporating synchronized adaptive density control to ensure high-quality 4D reconstruction. 

The formulation of our deformation model is detailed in Sec.\ref{subsec:deformation}, followed by an explanation of the initialization process in Sec.\ref{subsec:init}, and finally, the optimization details are discussed in Sec.~\ref{subsec:optimization}.

\subsection{Deformation Field}
\label{subsec:deformation}
How to model the deformation across frames is a key question for dynamic scene reconstruction of a monocular sequence. We adopt a similar strategy as in~\cite{yang_deformable_2023}, where a deformation field will be learned jointly with trainable 3D Gaussians. 

Unlike \cite{yang_deformable_2023}, we adopt multiple frames as reference frames to overcome the issue that some Gaussians may be invisible in the single reference frame, which usually leads to terrible results for high-deformation motions, which are quite common for human-centric scene. Specifically, we select $B$ frames of time $t_1,\dots,t_B$ as our reference frames. We use $\overline{\bs{x}}\in\mathbb{R}^{B\times 3}$ to denote the collection of key point locations $\bs{x}$ in the $B$ reference frames. Given time $t$ and location $\bs{x}$, the deformation field is defined as 
\begin{equation}
    (\bs{w},\delta\overline{\bs{x}},\delta\overline{\bs{r}},\delta\overline{\bs{s}}) = \mathcal{F}_\theta\big(\gamma(\textup{sg}(\overline{\bs{x}})),\gamma(t)\big)
\end{equation}
where $\textup{sg}$ is the stop-gradient operation, $\gamma(r):=\big(\sin(2^k\pi r),\cos(2^k\pi r)\big)_{k=0}^{L-1}$ denotes the positional encoding where $L$ is a hyperparameter and $\bs{w}\in\mathbb{R}^{B}$ represents the weights of reference frames and is subject to $\sum_{i=1}^Bw_i=1$.  $\delta\overline{\bs{x}},\delta\overline{\bs{r}},\delta\overline{\bs{s}}$ collect the offsets of location, rotation and scale, respectively, across reference frames. We then take the weighted sum of the offsets with weight vector $\bs{w}$ to get
\begin{equation}
\bs{x}_t^\star = \sum_{i=1}^Bw_{i}(\overline{\bs{x}}_{t_i}+\delta\overline{\bs{x}}_{t_i}).
\label{eq:deformation}
\end{equation}

$\bs{r}^\star_t,\bs{s}^\star_t$ are calculated in a similar way. Subsequently, we put the deformed 3D Gaussians $G(\bs{x}^\star_t,\bs{r}^\star_t,\bs{s}^\star_t)$ into the rendering pipeline to get the rendered image $\tilde{\bold I}_t$.

It is worth noting that some components in $\bs{x}$ may be unavailable due to invisible keypoints in some frames.
Further discussions on this minor issue and the details of network architecture are deferred to the appendix.

\subsection{Shape-aware Initialization}
\label{subsec:init}

Original 3D Gaussian splatting~\cite{kerbl_3d_2023} relies on the sparse points from Structure-from-Motion(SfM) which are typically obtained via COLMAP~\cite{schonberger_structure--motion_2016,schonberger_pixelwise_2016} to initialize all Gaussians locations. However, it is difficult to get a valid SfM results from a casual monocular video with significant human motions.  Actually, the sparse points belonging to human body may usually be completely missing in many scenarios. Thus, we propose a shape-aware initialization scheme for our deformation model.

\vspace{3pt}
\noindent\textbf{Alignment of depth maps.} When using estimated depth maps from off-the-shelf foundational models, two main challenges arise: (1) depth estimations are typically normalized, lacking consistency with world coordinates, and (2) general monocular depth estimations often perform poorly on human bodies due to limited training data, while human-specific models fail to capture comprehensive depth cues for the broader dynamic scene. To overcome these issues, we adopt a two-step approach to generate a high-quality aligned depth map, which provides a strong basis for initializing the deformation field.

The input data to generate the aligned depth map includes 3D sparse points ${\bs{q}_i}$, an estimated complete depth map~\cite{yang_depth_2024} $\mathbf{D}_{com}$, a human-specific depth map~\cite{khirodkar_sapiens_2024} $\mathbf{D}_{hum}$, and the corresponding human mask $\mathbf{M}$. First, we derive a sparse depth map $\mathbf{D}_{sparse}$ from ${\bs{q}_i}$. We align $\mathbf{D}_{com}$ to $\mathbf{D}_{sparse}$ by minimizing the following alignment loss:
$$
\mathcal{L}_{align} = \sum_{\bs{p}}\big|\big(s\cdot\mathbf{D}_{com}(\bs{p}) + t \big) - \mathbf{D}_{sparse}(\bs{p})\big|,
$$
optimizing for scale $s$ and shift $t$ using RANSAC~\cite{fischler_bolles_1981}. This produces an aligned map $\mathbf{D}_{com}^\star = s^\star\cdot \mathbf{D}_{com} + t^\star$. Next, we perform an affine transformation on $\mathbf{D}_{hum}$ to align it with the quantiles of $\mathbf{M}\circ \mathbf{D}_{com}^\star$. We used quantiles at 0.1, 0.5, and 0.9 for alignment, which resulted in the aligned human depth map $\mathbf{D}_{hum}^\star$. Our final aligned depth map is represented as:
$$\mathbf{D}^\star = \mathbf{M}\circ \mathbf{D}_{hum}^\star + (1-\mathbf{M})\circ \mathbf{D}_{com}^\star.$$

\noindent\textbf{Initialization of sparse Gaussians.} Using human UV maps $\bs{z}$, we are allowed to generate consistent 2D correspondences and frame-by-frame tracking. Specifically, we interpolate on the human UV maps to identify pixel coordinates for a list of predefined IUV keypoints. These 2D points are then lifted to 3D using the aligned depth map $\mathbf{D}^\star$. The initial rotation of each Gaussian is determined by aligning point clouds across frames using the orthogonal Procrustes method, while all initial scales are set to a constant. 

As for quasi-static background points, initialization is based on SfM points. Although the background may exhibit minor motion, the deformation field, without special initialization, can accommodate these dynamics well enough.

\vspace{3pt}
\noindent\textbf{Initialization of deformation field.}
Due to significant human motion and limited multi-view constraints, we initialize the deformation field using high-accuracy sparse 3D correspondences. We lift the aligned depth map from the previous step to 3D using known camera poses, then enforce alignment with the 3D points by minimizing the following deformation loss:
\vspace{-5pt}
\begin{align}
    \mathcal{L}_{deform} &:= \sum_{i=1}^B\sum_{t=1}^T\sum_{\bs{x}\in \mathcal{G}}\mathcal{M}_{t_i,t}(\bs{x})\Big(\lambda_1\|\bs{x}_{t_i}^\star - \overline{\bs{x}}_{t_i} \|_2^2 \nonumber\\
    &+ \lambda_2\|\bs{r}_{t_i}^\star - \overline{\bs{r}}_{t_i} \|_2^2 + \lambda_3\|\bs{s}_{t_i}^\star - \overline{\bs{s}}_{t_i} \|_2^2\Big)
    \label{Eq: L_deform}
\end{align}
where $\mathcal{G}$ collects all available Gaussians and $(\bs{x}_{t_i}^\star, \bs{r}_{t_i}^\star,\bs{s}_{t_i}^\star)$ are determined by our deformation field (\ref{eq:deformation}). The mask function $\mathcal{M}_{t_1,t_2}(\bs{x})$ is 1 if Gaussian $\bs{x}$ is visible in both $t_1$ and $t_2$, otherwise 0. Optimization is performed on the parameters of the deformation field $\theta$. 

After this step, we discard sparse Gaussians outside the reference frames and proceed to jointly optimize Gaussians within the reference frames and the deformation field.

\subsection{Optimization}
\label{subsec:optimization}



The primary objective is to optimize the deformation network by minimizing the discrepancy between the rendered images and the ground-truth images from the training dataset. This involves refining the alignment between the predicted scene geometry and the actual observations, ensuring more accurate and visually consistent reconstructions.

\noindent\textbf{Multi-frame density control.} Adaptive density control plays and important role in original 3DGS implementation~\cite{kerbl_3d_2023} to adaptively assign more Gaussians in detail-rich or high-frequency areas while prune redundant Gaussians in the plain areas. For deformation model with a single reference frame, the density control is straightforward. However, for our deformation model with multiple reference frames, we need to carefully apply the density control consistently to Gaussians over all frames.  Specifically, whenever we decide to clone or split one Gaussian, we replicate this operation across all frames with recalibrated orientations.

\noindent\textbf{Loss.} The loss term $\mathcal{L}$ is defined as the weighted sum of four components: L1 color loss $\mathcal{L}_{\textup{color}}$ and D-SSIM loss $\mathcal{L}_{\textup{D-SSIM}}$ comparing the rendered image $\tilde{\bold I}_t$ with the ground-truth image $\bold I_t$, depth loss $\mathcal{L}_{\textup{depth}}$ comparing the rendered depth map $\tilde{\mathbb{D}}$ with the aligned prior depth map $\mathbb{D}^\star$, and the rigidity loss $\mathcal{L}_{\textup{rigid}}$ is defined similar to \cite{wang_shape_2024}.

The final loss function is formulated as:
\begin{equation}
    \mathcal{L} = \mathcal{L}_{\textup{color}} + \mathcal{L}_{\textup{D-SSIM}} + \mathcal{L}_{\textup{depth}} + \mathcal{L}_{\textup{rigid}}.
\end{equation}

\vspace{1pt}
\noindent\textbf{Training strategy.} To avoid our geometry priors introduced in the initialization phase getting destructed too early due to the strong supervision from the loss, we adopt a warm-up strategy. In the initial epochs, an additional regularity loss term similar to Eq.~\ref{Eq: L_deform} is applied to ``freeze'' the Gaussian locations:
\begin{equation}
    \mathcal{L}_{freeze} = \zeta_{epoch}\cdot \sum_{t=1}^T\sum_{\bs{x}\in\mathcal{G}}\|\overline{\bs{x}}_{t} - 
    \textup{SRC}(\bs{x}_t^\star) \|_2^2
\label{Eq: freeze loss}
\end{equation}
where $\zeta_{epoch}=1-\frac{\min(epoch,N_{freeze})}{N_{freeze}}$ decreases as the epoch count grows. Here, $\textup{SRC}$ returns the source Gaussian corresponding to $\bs{x}_t$ that may have been splitted or cloned due to the density control strategy. After the first few epochs, $\mathcal{L}_{freeze}$ gradually decreases.

%% file: sec/sec-experiments.tex
\section{Experiments}
\input{tables/NeuMan}

\begin{figure*}[ht]
    \centering
    \includegraphics[width=\textwidth]{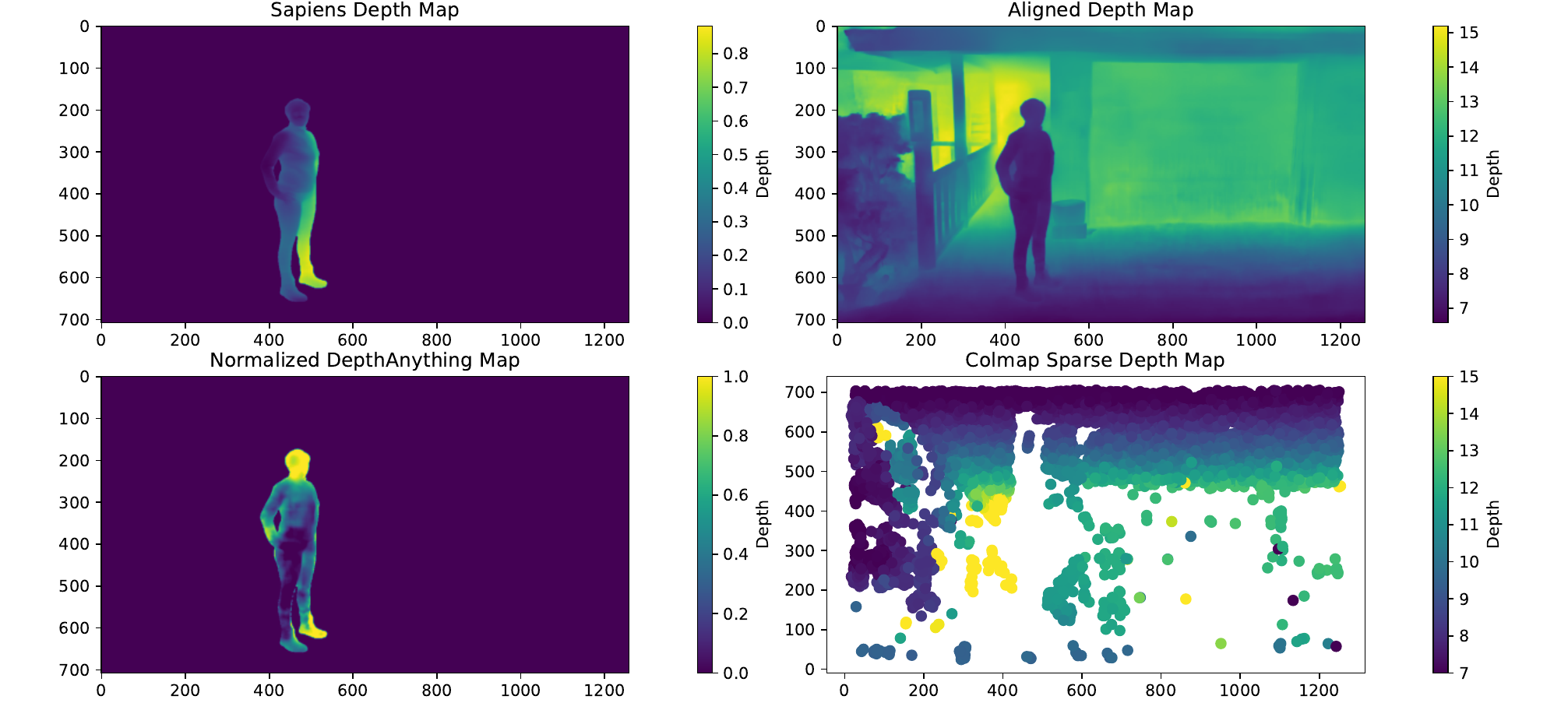}
    \vspace{-15pt}
    \caption{\textbf{Visualization of depth alignment.}\vspace{-20pt} }
    \label{fig:depth vis}
\end{figure*}

In this section, we assess our approach using real monocular datasets and conduct ablation studies to showcase its capability in reconstructing photo-realistic dynamic scenes featuring dramatic character action.

\noindent{\bf Datasets.} First, we assess our method on the widely utilized ZJU dataset~\cite{peng_neural_2021} under a monocular setting and demonstrate that it achieves state-of-the-art performance on objects suited to our model. Subsequently, we conducted quantitative, qualitative, and ablation studies on the more realistically captured NeuMan dataset~\cite{jiang_neuman_2022} to show that our method effectively provides adequate regularization for this under-constrained scenario, characterized by a camera moving around a human in motion.

\noindent{\bf Comparison Methods.}
We evaluate our methods against two classes of approaches: neural deformation methods and avatar fitting methods. The former category includes 4D Gaussian~\cite{wu_4d_2023} and Deformable Gaussian~\cite{yang_deformable_2023}. The latter category comprises GART~\cite{lei_gart_2023} and HUGS~\cite{kocabas_hugs_2023}. Shepe-of-Motion~\cite{wang_shape_2024} 

\noindent{\bf Implementation Details.}
We implement our method using PyTorch~\cite{paszke_pytorch_2019} and conduct the differentiable Gaussian rasterization with the gsplat framework~\cite{ye2024gsplatopensourcelibrarygaussian} which provides both rgb and depth rendering. For training, the first 5k iterations are used to optimize the deformation network, and the remaining iterations are used to optimize the 3D Gaussians in reference frames and deformation field jointly. For optimization, a single Adam optimizer~\cite{KingBa15} is used with different learning rates for the components. The whole training process including individual preparation takes approximately 40 minutes on a NVIDIA RTX 4090.

\subsection{Results}

\noindent{\bf Comparison on NeuMan dataset.~\cite{jiang_neuman_2022}}
Table~\ref{tab:neuman_human_scene} provides a comparative analysis of our method against existing methods on the NeuMan dataset. In particular, GART is not capable of modeling the environment other than the human. Hence we synthesize the complete novel view by combining the human rendering generated by GART and the background scene rendering provided by 4DGS in light of the human segmentation.

Our method consistently shows superior performance across all metrics and scenarios, highlighting its effectiveness in generating high-quality reconstructions from monocular recordings. In addition, the significant performance advantage of GART over 4DGS and Deformable-GS clearly indicates the critical importance of the SMPL prior.

\begin{figure}
    \centering
    \includegraphics[width=\columnwidth]{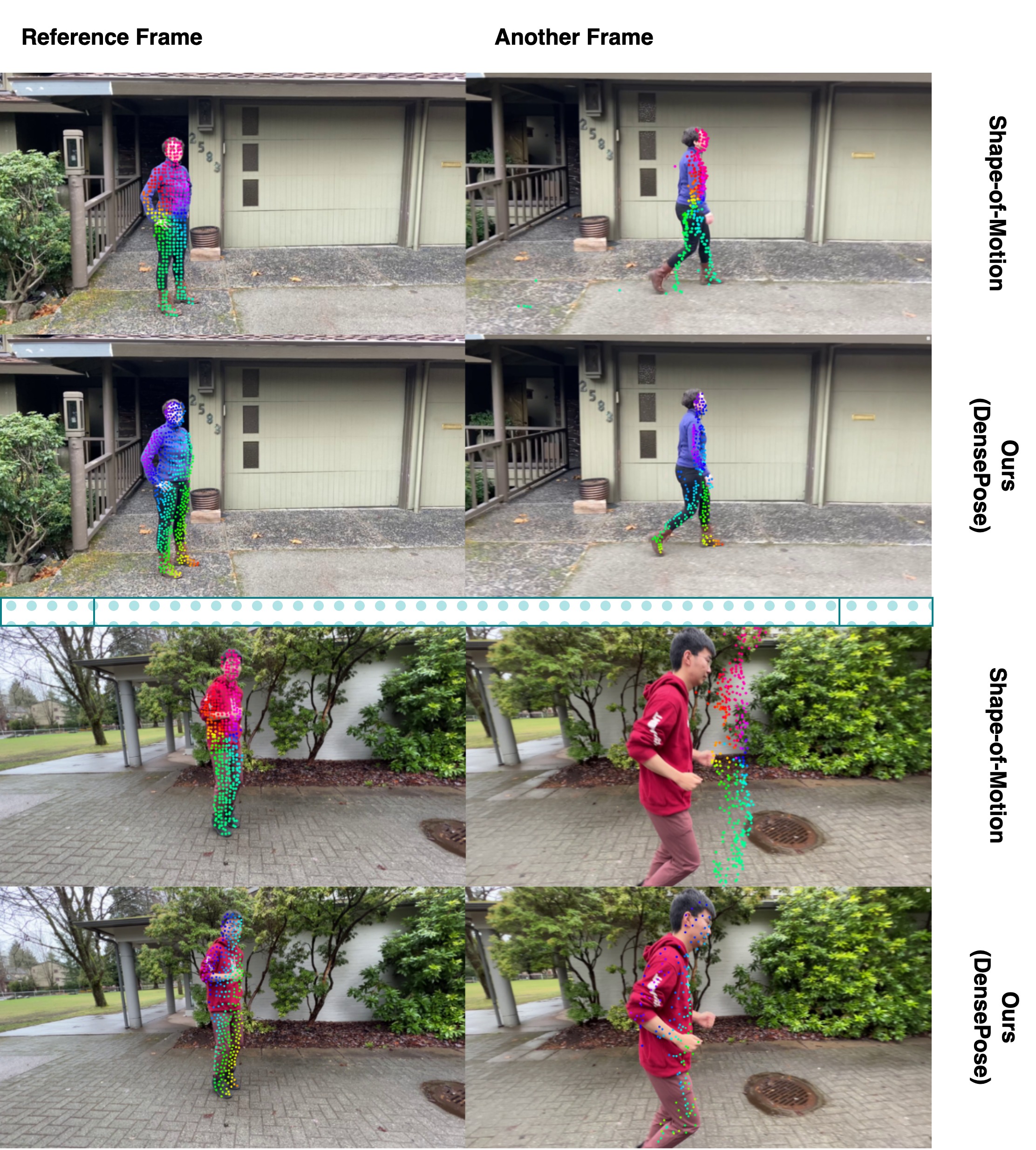}
    \caption{\footnotesize\textbf{Comparison of keypoint initializations.} The erroneous initialization of keypoints is the primary factor degrading the reconstruction performance of Shape-of-Motion~\cite{wang_shape_2024} on the NeuMan dataset. Note that, for our method, we visualize only the keypoints in the other frame that remain visible in the reference frame.}
    \label{fig:init_comp}
\end{figure}

\begin{figure*}[!ht]
\centering
\includegraphics[width=\textwidth]{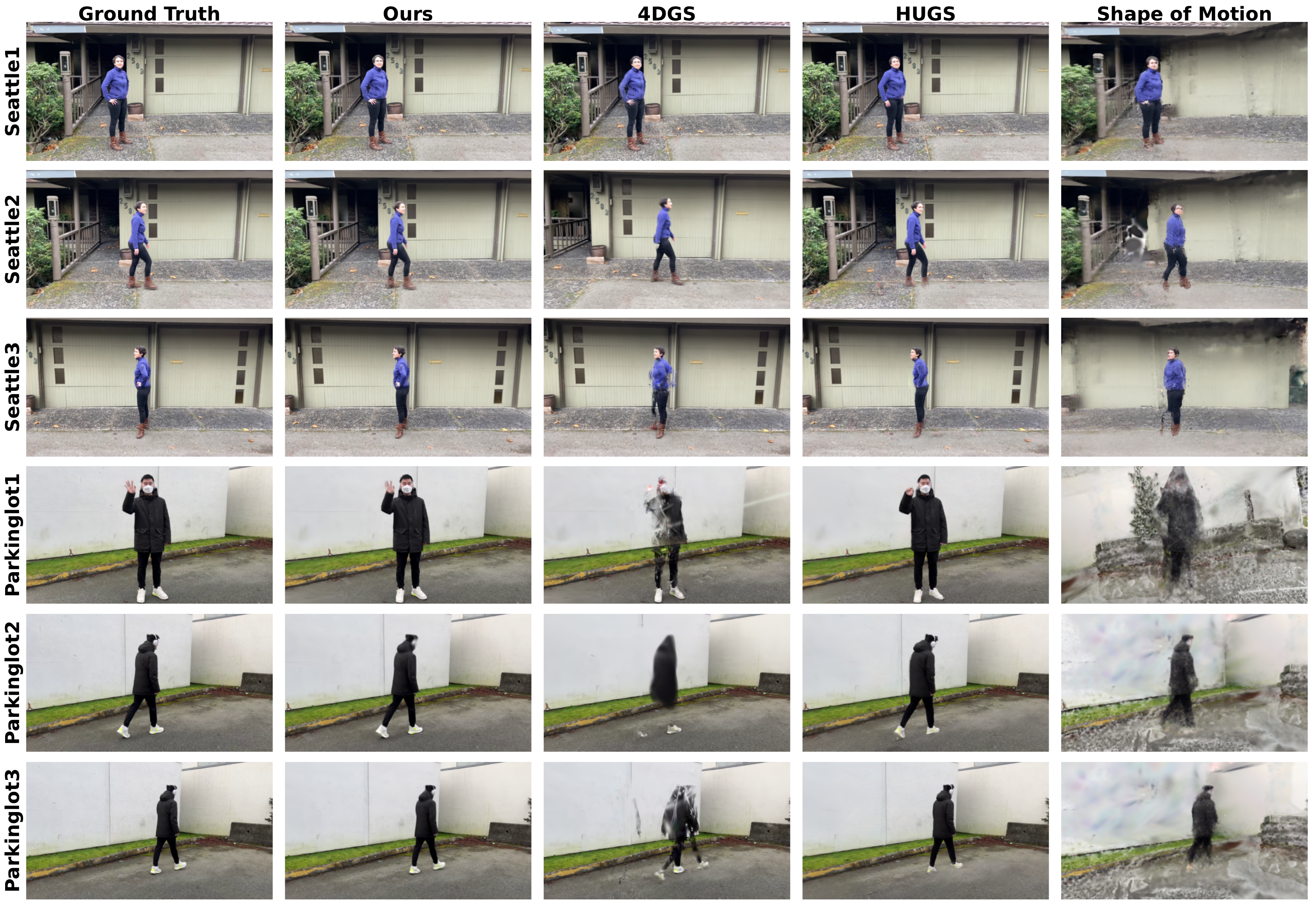}
\caption{\textbf{Qualitative comparison of baselines and our method on real dataset of casual monocular videos.}}
\label{fig:neuman}
\end{figure*}

Figure~\ref{fig:neuman} presents a qualitative comparison of our method against Deformable 3DGS~\cite{yang_deformable_2023}, 4DGS~\cite{yang_real-time_2024}, which lacks a human shape prior, HUGS~\cite{kocabas_hugs_2023}, which leverages a strong human template, and Shape-of-Motion~\cite{wang_shape_2024}, which employs general data-driven priors on the NeuMan dataset. For Shape-of-Motion, we follow the authors’ official instructions for custom sequences.

Generally, Deformable 3DGS and 4DGS yield significantly lower-quality renderings from novel viewpoints and, in some cases, fail to capture the human figure in the scene due to missing geometric information. Although Shape-of-Motion integrates data-driven priors such as depth maps and long-range 2D keypoints, it performs poorly because the generic 2D keypoint correspondence algorithm, such as BootsTAP~\cite{doersch2024bootstapbootstrappedtrainingtrackinganypoint} that was adopted in Shape-of-Motion, fails to track correct trajectories of human. As shown in Figure~\ref{fig:init_comp}, its 2D keypoint tracking quickly diverges from the reference frame, whereas our method which derives 2D keypoints from DensePose stably tracks human motion.  While HUGS~\cite{kocabas_hugs_2023} performs better due to its human template, it often produces inaccurate renderings of human appearance and poses due to errors in 3D pose estimation. In contrast, our method consistently delivers high-quality reconstructions by incorporating high-quality 2D visual cues, showcasing its robustness and effectiveness in managing complex human motions.

\vspace{5pt}
\noindent{\bf Comparison on ZJU Mocap dataset.~\cite{peng_neural_2021}}
Table~\ref{tab:zju} compares the performance of our method with other state-of-the-art techniques, namely 4DGS, Deformable-GS, HUGS*, and GART, on test images from the ZJU Mocap dataset. The evaluation focuses on various scenarios identified by their dataset numbers: 377, 386, 387, 392, 393, and 394. We adopt the same camera view settings as \cite{peng_animatable_2021} for training and testing. Our method consistently demonstrates superior performance in terms of PSNR and SSIM in all scenarios. In the context of the LPIPS metric, our method generally achieves competitive results, sometimes outperforming others.

\vspace{3pt}
\noindent{\bf Depth alignment visualization.} Figure~\ref{fig:depth vis} highlights the importance of incorporating a human-specific depth map to ensure that our model begins with an initialization featuring accurate geometry. Generic depth estimation models~\cite{yang_depth_2024, godard2017unsupervisedmonoculardepthestimation} often do not produce an ordinarily consistent depth for humans. Although recent methods like MoDGS~\cite{liu2024modgsdynamicgaussiansplatting} address this issue by introducing additional realignment steps to mitigate this issue, we empirically observe that combining a generic depth model with a human-specific depth model achieves significantly better results in human-centric scenes while maintaining simplicity.

\subsection{Ablation Study}
As shown in Table~\ref{tab:ablation}, our ablation studies evaluated the effect of removing several key components on reconstruction performance, using the real-world NeuMan data set for evaluation. The first component investigated was the role of human-centric depth estimation. For this, we tested initializing and supervising our model using only a generic depth map~\cite{yang_depth_2024}. The second component examined the impact of multiple reference frames by comparing it to the traditional deformation model, which is based on a single canonical frame. Lastly, we tested the necessity of initializing the deformation field using a randomly sampled deformation field, as in most previous approaches. Table~\ref{tab:ablation} highlights that each of these components significantly improves the performance of the model, underscoring their importance in achieving high-quality reconstructions.

\input{tables/ablation}
\input{tables/ZJU}

%% file: tables/NeuMan.tex
\definecolor{tabfirst}{rgb}{1, 0.7, 0.7}
\definecolor{tabsecond}{rgb}{1, 0.85, 0.7}
\definecolor{tabthird}{rgb}{1, 1, 0.7}

\begin{table*}[h]
    \centering
    \resizebox{\textwidth}{!}{
    \begin{tabular}{c|ccc|ccc|ccc}
    \toprule
        & \multicolumn{3}{c|}{\textbf{Seattle}} & \multicolumn{3}{c|}{\textbf{Citron}} & \multicolumn{3}{c}{\textbf{Parking}} \\
    \midrule
        & PSNR $\uparrow$ & SSIM $\uparrow$ & LPIPS* $\downarrow$ & PSNR $\uparrow$ & SSIM $\uparrow$ & LPIPS* $\downarrow$ & PSNR $\uparrow$ & SSIM $\uparrow$ & LPIPS* $\downarrow$  \\
    \midrule
    4DGS & 21.03 & 0.75 & 161.1 & 20.34 & 0.88 & 214.5 & 22.96 & 0.84 & 234.1 \\
Deformable-GS & 20.39 & 0.88 & 169.4 & 19.87 & 0.87 & 241.6 & 23.11 & 0.87 & 251.2 \\
    HUGS &  25.94 &  0.85 & 130 &  25.54 &  0.86 &  150 &  26.86 &  0.85 &  220 \\
    Shape-of-Motion &  19.06 &  0.78 &  185.7 &  21.78 &  0.87 &  221.4 &  18.72 &  0.79 &  331.4 \\
    \midrule
  Ours &  \cellcolor{tabfirst}27.68 & \cellcolor{tabfirst} 0.90 & \cellcolor{tabfirst} 106.4 & \cellcolor{tabfirst} 26.51 & \cellcolor{tabfirst} 0.91 & \cellcolor{tabfirst} 126.4 & \cellcolor{tabfirst} 27.15 & \cellcolor{tabfirst} 0.89 & \cellcolor{tabfirst} 168.2 \\
    \midrule\midrule
            & \multicolumn{3}{c|}{\textbf{Bike}} & \multicolumn{3}{c|}{\textbf{Jogging}} & \multicolumn{3}{c}{\textbf{Lab}} \\
    \midrule
    4DGS & 21.39 & 0.81 & 182.3 & 17.19 & 0.75 & 153.2 & 22.34 & 0.90 & 134.5 \\
    Deformable-GS & 20.17 & 0.80 & 172.9 & 18.23 & 0.79 & 191.2 & 21.69 & 0.91 & 129.9 \\

    HUGS &  25.46 &  0.84 & 130 &  23.75 &  0.78 &  220 &  26.00 &  \cellcolor{tabfirst}0.92 & 90 \\
    Shape-of-Motion &  19.47 &  0.77 &  174.8 &  21.19 &  0.79 &  204.6 &  21.84 &  0.88 &  128.7 \\
    
    \midrule
     Ours &   \cellcolor{tabfirst} 26.31 & \cellcolor{tabfirst}0.97 &  \cellcolor{tabfirst}98.1 & \cellcolor{tabfirst} 24.49 & \cellcolor{tabfirst}0.81 &          \cellcolor{tabfirst} 124.8 & \cellcolor{tabfirst} 27.48 & 0.91 &  \cellcolor{tabfirst}79\\
    \bottomrule
    \end{tabular}  
    }
    \caption{\footnotesize{Comparison of ours method with previous work on test images of the NeuMan dataset~\cite{jiang_neuman_2022} using PSNR, SSIM and 1000x LPIPS metrics.}}
    \label{tab:neuman_human_scene}
\end{table*}

%% file: tables/ablation.tex
\begin{table}[t]
    \centering
    \begin{tabular}{c|ccc}
    \toprule
        & PSNR $\uparrow$ & SSIM $\uparrow$ & LPIPS $\downarrow$  \\
    \midrule
    w/o HDM & 24.18 & 0.864 & 0.159 \\

    w/o MF &  25.95 &  0.883 &  0.141 \\
    w/o DFI & 22.34 & 0.803 & 0.214 \\
    \midrule
    Complete & \cellcolor{tabfirst} 26.60 & \cellcolor{tabfirst} 0.898 & \cellcolor{tabfirst} 0.117 \\ 

    \bottomrule
    \end{tabular}  
    
\caption{\footnotesize \textbf{Ablation study.} The performance is evaluated over full images using PSNR, SSIM and LPIPS metrics on NeuMan dataset. \textbf{HDM}: Human-only depth map, \textbf{MF}: Multiple reference frames, \textbf{DFI}: Deformation field initialization \vspace{-15pt}}
\label{tab:ablation}
\end{table}

%% file: tables/ZJU.tex
\definecolor{tabfirst}{rgb}{1, 0.7, 0.7}
\definecolor{tabsecond}{rgb}{1, 0.85, 0.7}
\definecolor{tabthird}{rgb}{1, 1, 0.7}

\begin{table}[t]
    \centering
    \begin{tabular}{c|ccc}
    \toprule
        & PSNR $\uparrow$ & SSIM $\uparrow$ & LPIPS*\\
    \midrule
    4DGS & 27.18 & 0.84 & 97.2 \\
    Deformable-GS & 28.49 & 0.86 & 97.7 \\

    HUGS* &  30.80 &  0.98 &  20  \\
    Shape-of-Motion & 26.87 & 0.82 & 102.4  \\
 GART & \cellcolor{tabfirst} 31.90 & \cellcolor{tabfirst} 0.97 &  \cellcolor{tabsecond} 18.8 \\
    \midrule
  Ours &  \cellcolor{tabsecond}30.92 & \cellcolor{tabsecond} 0.96 & \cellcolor{tabfirst} 18.9 \\
    \bottomrule
    \end{tabular}
    \caption{\footnotesize Comparison of ours method with previous work on test images of the ZJU Mocap dataset~\cite{peng_neural_2021} using PSNR, SSIM and 1000x LPIPS metrics. The evaluation results of HUGS are quoted directly from \cite{kocabas_hugs_2023}, and suffer from low precision for LPIPS. \vspace{-15pt} }
    \label{tab:zju}
\end{table}

%% file: sec/sec-conclusion.tex
\subsection{Limitations}
While our method can produce high-quality dynamic reconstructions from monocular videos featuring fast human motion, it faces several challenges and limitations. 
Firstly, our approach relies on the semantic priors provided by an external model. Moreover a set of accurate camera parameters have to be given and our reconstruction is highly sensitive to the errors of camera poses. Secondly, unlike most template-based human reconstruction work, our method cannot accommodate renderings with novel human poses. Lastly, designing a more complex model is necessary to effectively handle scenarios involving multiple people, particularly in cases with occlusions.

\section{Conclusion}

In conclusion, we present \emph{ShapeGaussian}, a novel approach to capture high-fidelity dynamic scene reconstruction of human motion captured with a monocular camera. This method achieves robust and fast reconstruction while enhancing rendering quality, as validated by quantitative metrics. Unlike template-based methods, it is free from template restriction and artifacts caused by inaccurate pose estimation, ensuring more natural and adaptable results. Technically, we introduced a novel strategy that employs multiple reference frames to effectively track invisible Gaussians in a monocular setting, overcoming key challenges in dynamic scene reconstruction.

%% file: sec/sec-appendix.tex
\appendix
The appendix provides more implementation details and additional experimental results to support the main paper. 

\begin{table*}[ht]
    \centering
    \begin{tabular}{|c|c|c|c|c|c|c|}
    \hline
     & Bike & Citron & Jogging & Lab &  Parkinglot & Seattle\\
    \hline 
    SOM & 25.27 & 24.95 & 24.05 & 25.94 & 25.78 & 26.31 \\
    \hline 
    Ours & 26.31 & 26.51 & 24.49 & 27.48 & 27.15 & 27.68 \\
    \hline
    \end{tabular}
    \caption{Comparison of PSNR between our method and Shape-of-Motion~\cite{wang_shape_2024}.}
\end{table*}

\section{Implementation Details}
\label{sec-app:implementation}
This section outlines the implementation specifics of the deformation model employed in our method, providing clarity on network design, training strategies, and practical considerations.

\subsection{Deformation Model}
\begin{figure}
    \centering
    \includegraphics[width=0.5\columnwidth]{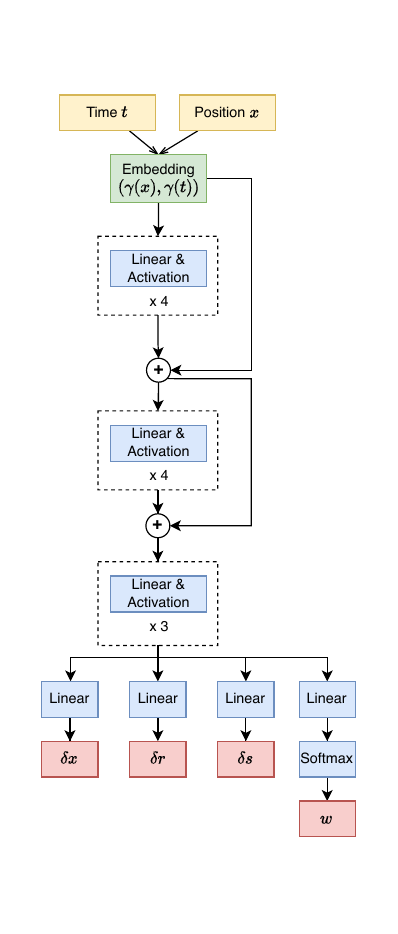}
    \caption{The network architecture for modeling deformation.}
    \label{fig:network_architecture}
\end{figure}

The deformation model is based on a neural network that processes Gaussian features and predicts transformations across frames. Figure~\ref{fig:network_architecture} illustrates our network architecture, which builds upon \cite{yang_deformable_2023}. 
The embedding function $\gamma$ enriches the input features by encoding spatial frequencies: $\gamma$ is defined by
\begin{equation}
    \gamma(p) := (\sin(2^k\pi p), \cos(2^k \pi p))_{k=0}^{L-1},
\end{equation} 
where $L=10$. An additional output channel predicts weights for each reference frame, enabling flexible and accurate deformation modeling.

During initialization, we align the deformation model with semantic priors provided by vision foundation models. The loss function defined by
\begin{align}
    \mathcal{L}_{deform} &:= \sum_{i=1}^B\sum_{t=1}^T\sum_{\bs{x}\in \mathcal{G}}\mathcal{M}_{t_i,t}(\bs{x})\Big(\lambda_1\|\bs{x}_{t_i}^\star - \overline{\bs{x}}_{t_i} \|_2^2 \nonumber\\
    &+ \lambda_2\|\bs{r}_{t_i}^\star - \overline{\bs{r}}_{t_i} \|_2^2 + \lambda_3\|\bs{s}_{t_i}^\star - \overline{\bs{s}}_{t_i} \|_2^2\Big),
\end{align}
ensures consistency between the predicted and reference semantic priors.

After initialization, Gaussians deform according to:
$$
\bs{x}_t^\star:=\sum_{i=1}^B w_i(\bar{\bs{x}}_{t_i}+\delta\bar{\bs{x}}_{t_i}),
$$
where $\delta\bar{\bs{x}}_{t_i}$ and $w_i$ are network outputs. For Gaussians that are not visible in certain frames, we initialize them as the mean of visible Gaussians and enforce corresponding $w_i$ towards 0 during optimization using the regularization term
$$
\mathcal{L}_{weight} := \sum_{i=1}^B\sum_{t=1}^T\sum_{\bs{x}\in \mathcal{G}}\Big(1-\mathcal{M}_{t_i,t}(\bs{x})\Big)\|w_i(\bs{x},t)\|_2^2.
$$
This regularization penalizes weights for invisible Gaussians, ensuring robustness in scenarios with partial visibility.

\vspace{6pt}
\noindent\textbf{Selection of reference frames.} In our method, reference frames are selected based on two criteria: (1) the union of adjacent reference frames should maximize the coverage of keypoints, and (2) the reference frames should be evenly distributed across the video. Given a predefined parameter  B , representing the number of reference frames, we empirically optimize these criteria by minimizing the following cost function:
\begin{align}
    \mathcal{L}_{ref} := &\Var(\{t_{j+1}-t_j \mid j=1,\dots,B-1\}) / T \nonumber\\
            & - \frac{\lambda_{ref}}{|\mathcal{P}|} \sum_{i} \Big|\bigcup_{i\leq j<i+N_{neigh}}\mathcal{P}_{t_j}\Big| ,
            \label{eq:find_ref}
\end{align}
where \( \Var \) computes the variance of time intervals between adjacent reference frames, $\mathcal{P}_{t_j}$ denotes the set of keypoints visible in the  $j$-th reference frame at timestamp $t_j$, and $\mathcal{P}$ is the set of all keypoints. The parameter $\lambda_{ref}$ controls the trade-off between the two criteria. 

In our experiments, we set  $B=4, \lambda_{ref} = 0.2$ and  $N_{neigh} = 3$ . Since the number of reference frames is relatively small, an exhaustive search is computationally feasible to identify the optimal reference frames that minimize $\mathcal{L}_{ref}$  as defined in Eq.~\ref{eq:find_ref}.


\subsection{Synchronized Density Control}
Our method faces two primary challenges due to the use of multiple reference frames: (1) ensuring consistent density control during training to maintain the coherence of Gaussians across frames, and (2) managing the potentially large number of Gaussians, which scales with the number of reference frames. To address the second challenge, we freeze all background Gaussians and dynamically update only those associated with the human subject, significantly reducing computational overhead.

To preserve geometric consistency, when a Gaussian in one reference frame is split or cloned, we identify its k-nearest neighbors across all reference frames. These neighbors are analyzed to compute relative rotation and scaling factors, which are then uniformly applied to the corresponding Gaussians in all reference frames.

\section{Comparison with Generic Long-Term Tracking Methods}
\label{sec-app:more_results}

In this section, we detail why our method surpasses existing state-of-the-art (SOTA) methods~\cite{wang_shape_2024,liu2024modgsdynamicgaussiansplatting,lei2024moscadynamicgaussianfusion}, which rely on general long-term tracking priors. Specifically, we compare the qualitative and quantitative results of our approach with Shape-of-Motion~\cite{wang_shape_2024}, leveraging its publicly available source code. We processed casual videos following the official repository’s guidelines.

\begin{figure}
\centering
\includegraphics[width=0.8\columnwidth]{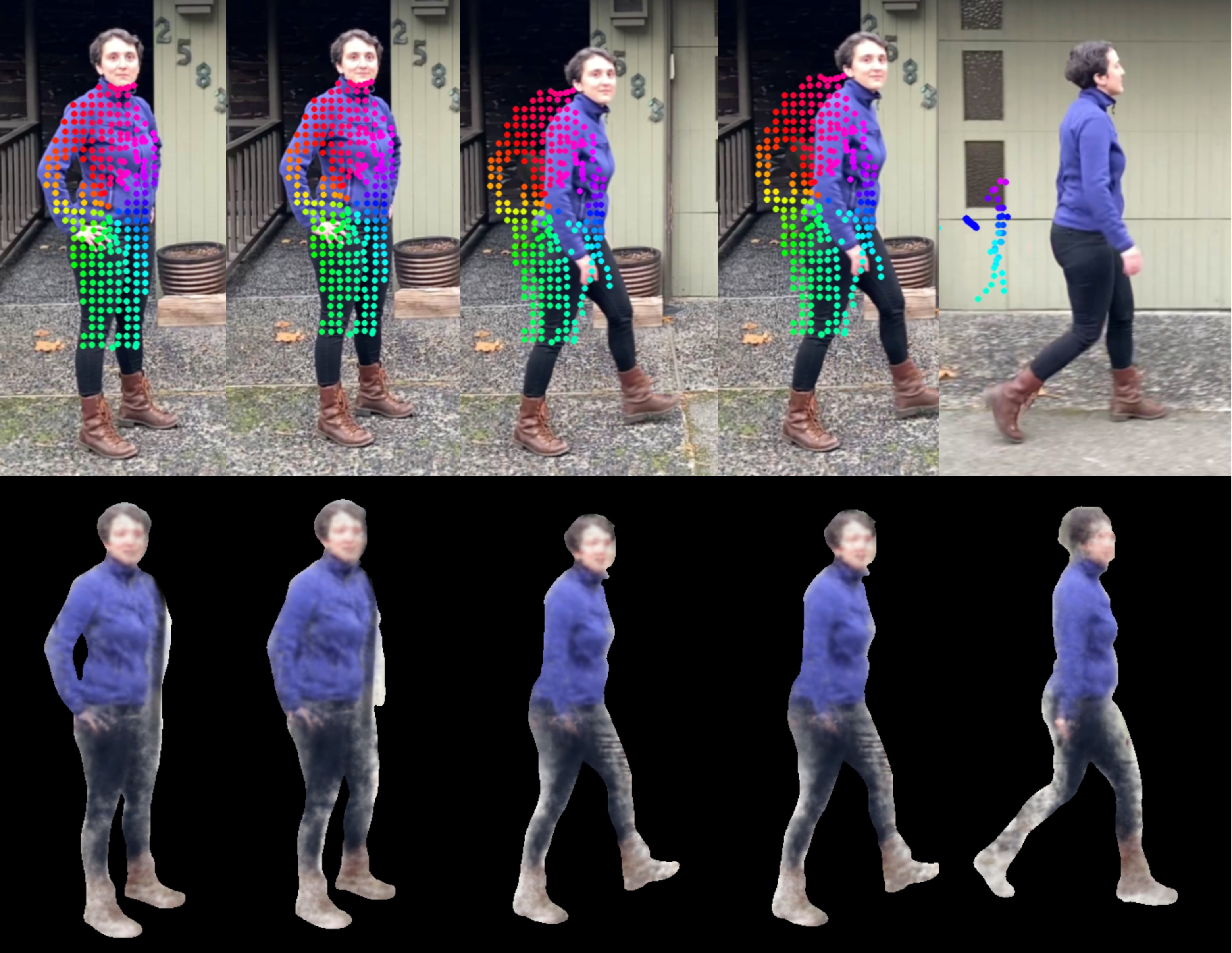}
\caption{Shape-of-motion fails to reconstruct highly-deformable human motion due to lack of human shape priors.}
\label{fig:som}
\end{figure}

Figure~\ref{fig:som} illustrates the limitations of Shape-of-Motion in reconstructing human-centric casual videos. The top row shows 2D keypoint tracking results, while the bottom row depicts 3D reconstructions on interpolated frames. The outputs are notably noisy and inconsistent due to the reliance on generic long-range tracking models like TAPIR~\cite{doersch_tapir_2023}, which lack human-specific priors. This results in frequent keypoint loss in scenarios involving fast-moving human motions and dynamic camera movements, both of which are common in casual video settings.

In contrast, our method effectively handles these challenges by incorporating human-specific priors and leveraging multiple reference frames, resulting in more robust and accurate reconstructions.

%% file: main.bib
@String(CVPR= {IEEE Conf. Comput. Vis. Pattern Recog.})

@String(ECCV= {Eur. Conf. Comput. Vis.})

@String(ICLR = {Int. Conf. Learn. Represent.})

@String(IJCAI = {IJCAI})

@String(CVPR  = {CVPR})

@String(ECCV  = {ECCV})

@String(ICLR  = {ICLR})


%% file: references.bib
@misc{yang_depth_2024,
	title = {Depth {Anything} {V2}},
	url = {http://arxiv.org/abs/2406.09414},
	doi = {10.48550/arXiv.2406.09414},
	urldate = {2024-11-12},
	publisher = {arXiv},
	author = {Yang, Lihe and Kang, Bingyi and Huang, Zilong and Zhao, Zhen and Xu, Xiaogang and Feng, Jiashi and Zhao, Hengshuang},
	month = oct,
	year = {2024},
	note = {arXiv:2406.09414},
}

@misc{wang_shape_2024,
	title = {Shape of {Motion}: {4D} {Reconstruction} from a {Single} {Video}},
	shorttitle = {Shape of {Motion}},
	url = {http://arxiv.org/abs/2407.13764},
	doi = {10.48550/arXiv.2407.13764},
	urldate = {2024-11-10},
	publisher = {arXiv},
	author = {Wang, Qianqian and Ye, Vickie and Gao, Hang and Austin, Jake and Li, Zhengqi and Kanazawa, Angjoo},
	month = jul,
	year = {2024},
	note = {arXiv:2407.13764},
}

@misc{cao_openpose_2019,
	title = {{OpenPose}: {Realtime} {Multi}-{Person} {2D} {Pose} {Estimation} using {Part} {Affinity} {Fields}},
	shorttitle = {{OpenPose}},
	url = {http://arxiv.org/abs/1812.08008},
	doi = {10.48550/arXiv.1812.08008},
	urldate = {2024-11-09},
	publisher = {arXiv},
	author = {Cao, Zhe and Hidalgo, Gines and Simon, Tomas and Wei, Shih-En and Sheikh, Yaser},
	month = may,
	year = {2019},
	note = {arXiv:1812.08008},
}

@misc{jiang_instantavatar_2022,
	title = {{InstantAvatar}: {Learning} {Avatars} from {Monocular} {Video} in 60 {Seconds}},
	shorttitle = {{InstantAvatar}},
	url = {http://arxiv.org/abs/2212.10550},
	urldate = {2024-11-09},
	publisher = {arXiv},
	author = {Jiang, Tianjian and Chen, Xu and Song, Jie and Hilliges, Otmar},
	month = dec,
	year = {2022},
	note = {arXiv:2212.10550},
}

@misc{guler_densepose_2018,
	title = {{DensePose}: {Dense} {Human} {Pose} {Estimation} {In} {The} {Wild}},
	shorttitle = {{DensePose}},
	url = {http://arxiv.org/abs/1802.00434},
	doi = {10.48550/arXiv.1802.00434},
	urldate = {2024-11-09},
	publisher = {arXiv},
	author = {Güler, Rıza Alp and Neverova, Natalia and Kokkinos, Iasonas},
	month = feb,
	year = {2018},
	note = {arXiv:1802.00434},
}

@misc{wang_tracking_2023,
	title = {Tracking {Everything} {Everywhere} {All} at {Once}},
	url = {http://arxiv.org/abs/2306.05422},
	doi = {10.48550/arXiv.2306.05422},
	urldate = {2024-11-09},
	publisher = {arXiv},
	author = {Wang, Qianqian and Chang, Yen-Yu and Cai, Ruojin and Li, Zhengqi and Hariharan, Bharath and Holynski, Aleksander and Snavely, Noah},
	month = sep,
	year = {2023},
	note = {arXiv:2306.05422},
}

@misc{neoral_mft_2023,
	title = {{MFT}: {Long}-{Term} {Tracking} of {Every} {Pixel}},
	shorttitle = {{MFT}},
	url = {http://arxiv.org/abs/2305.12998},
	doi = {10.48550/arXiv.2305.12998},
	urldate = {2024-11-09},
	publisher = {arXiv},
	author = {Neoral, Michal and Šerých, Jonáš and Matas, Jiří},
	month = nov,
	year = {2023},
	note = {arXiv:2305.12998},
}

@misc{karaev_cotracker_2024,
	title = {{CoTracker}: {It} is {Better} to {Track} {Together}},
	shorttitle = {{CoTracker}},
	url = {http://arxiv.org/abs/2307.07635},
	doi = {10.48550/arXiv.2307.07635},
	urldate = {2024-11-09},
	publisher = {arXiv},
	author = {Karaev, Nikita and Rocco, Ignacio and Graham, Benjamin and Neverova, Natalia and Vedaldi, Andrea and Rupprecht, Christian},
	month = oct,
	year = {2024},
	note = {arXiv:2307.07635},
}

@misc{harley_particle_2022,
	title = {Particle {Video} {Revisited}: {Tracking} {Through} {Occlusions} {Using} {Point} {Trajectories}},
	shorttitle = {Particle {Video} {Revisited}},
	url = {http://arxiv.org/abs/2204.04153},
	doi = {10.48550/arXiv.2204.04153},
	urldate = {2024-11-09},
	publisher = {arXiv},
	author = {Harley, Adam W. and Fang, Zhaoyuan and Fragkiadaki, Katerina},
	month = jul,
	year = {2022},
	note = {arXiv:2204.04153},
}

@misc{doersch_tapir_2023,
	title = {{TAPIR}: {Tracking} {Any} {Point} with per-frame {Initialization} and temporal {Refinement}},
	shorttitle = {{TAPIR}},
	url = {http://arxiv.org/abs/2306.08637},
	doi = {10.48550/arXiv.2306.08637},
	urldate = {2024-11-09},
	publisher = {arXiv},
	author = {Doersch, Carl and Yang, Yi and Vecerik, Mel and Gokay, Dilara and Gupta, Ankush and Aytar, Yusuf and Carreira, Joao and Zisserman, Andrew},
	month = aug,
	year = {2023},
	note = {arXiv:2306.08637},
}

@inproceedings{xu_gmflow_2022,
	address = {New Orleans, LA, USA},
	title = {{GMFlow}: {Learning} {Optical} {Flow} via {Global} {Matching}},
	copyright = {https://doi.org/10.15223/policy-029},
	isbn = {978-1-66546-946-3},
	shorttitle = {{GMFlow}},
	url = {https://ieeexplore.ieee.org/document/9879713/},
	doi = {10.1109/CVPR52688.2022.00795},
	language = {en},
	urldate = {2024-11-09},
	booktitle = {2022 {IEEE}/{CVF} {Conference} on {Computer} {Vision} and {Pattern} {Recognition} ({CVPR})},
	publisher = {IEEE},
	author = {Xu, Haofei and Zhang, Jing and Cai, Jianfei and Rezatofighi, Hamid and Tao, Dacheng},
	month = jun,
	year = {2022},
	pages = {8111--8120},
}

@misc{teed_raft_2020,
	title = {{RAFT}: {Recurrent} {All}-{Pairs} {Field} {Transforms} for {Optical} {Flow}},
	shorttitle = {{RAFT}},
	url = {http://arxiv.org/abs/2003.12039},
	doi = {10.48550/arXiv.2003.12039},
	urldate = {2024-11-09},
	publisher = {arXiv},
	author = {Teed, Zachary and Deng, Jia},
	month = aug,
	year = {2020},
	note = {arXiv:2003.12039},
}

@misc{sun_pwc-net_2018,
	title = {{PWC}-{Net}: {CNNs} for {Optical} {Flow} {Using} {Pyramid}, {Warping}, and {Cost} {Volume}},
	shorttitle = {{PWC}-{Net}},
	url = {http://arxiv.org/abs/1709.02371},
	doi = {10.48550/arXiv.1709.02371},
	urldate = {2024-11-09},
	publisher = {arXiv},
	author = {Sun, Deqing and Yang, Xiaodong and Liu, Ming-Yu and Kautz, Jan},
	month = jun,
	year = {2018},
	note = {arXiv:1709.02371},
}

@misc{shi_videoflow_2023,
	title = {{VideoFlow}: {Exploiting} {Temporal} {Cues} for {Multi}-frame {Optical} {Flow} {Estimation}},
	shorttitle = {{VideoFlow}},
	url = {https://openaccess.thecvf.com/content/ICCV2023/html/Shi_VideoFlow_Exploiting_Temporal_Cues_for_Multi-frame_Optical_Flow_Estimation_ICCV_2023_paper.html},
	language = {en},
	urldate = {2024-11-09},
	author = {Shi, Xiaoyu and Huang, Zhaoyang and Bian, Weikang and Li, Dasong and Zhang, Manyuan and Cheung, Ka Chun and See, Simon and Qin, Hongwei and Dai, Jifeng and Li, Hongsheng},
	year = {2023},
	pages = {12469--12480},
}

@inproceedings{lucas_iterative_1981,
	address = {Vancouver, Canada},
	title = {An {Iterative} {Image} {Registration} {Technique} with an {Application} to {Stereo} {Vision}},
	volume = {2},
	url = {https://hal.science/hal-03697340},
	urldate = {2024-11-09},
	booktitle = {{IJCAI}'81: 7th international joint conference on {Artificial} intelligence},
	author = {Lucas, Bruce D and Kanade, Takeo},
	month = aug,
	year = {1981},
	pages = {674--679},
}

@article{fischler_bolles_1981,
  author = {Fischler, M. and Bolles, R.},
  journal = { Communications of the ACM},
  number = { 6},
  pages = { 381-395},
  title = { Random Sample Consensus: A Paradigm for Model Fitting with Applications to Image Analysis and Automated Cartography},
  url = {/brokenurl# http://publication.wilsonwong.me/load.php?id=233282275},
  volume = { 24},
  year = { 1981}
}

@article{horn_determining_1981,
	title = {Determining optical flow},
	volume = {17},
	issn = {0004-3702},
	url = {https://www.sciencedirect.com/science/article/pii/0004370281900242},
	doi = {10.1016/0004-3702(81)90024-2},
	number = {1},
	urldate = {2024-11-09},
	journal = {Artificial Intelligence},
	author = {Horn, Berthold K. P. and Schunck, Brian G.},
	month = aug,
	year = {1981},
	pages = {185--203},
}

@inproceedings{brox_high_2004,
	address = {Berlin, Heidelberg},
	title = {High {Accuracy} {Optical} {Flow} {Estimation} {Based} on a {Theory} for {Warping}},
	isbn = {978-3-540-24673-2},
	doi = {10.1007/978-3-540-24673-2_3},
	language = {en},
	booktitle = {Computer {Vision} - {ECCV} 2004},
	publisher = {Springer},
	author = {Brox, Thomas and Bruhn, Andrés and Papenberg, Nils and Weickert, Joachim},
	editor = {Pajdla, Tomás and Matas, Jiří},
	year = {2004},
	pages = {25--36},
}

@inproceedings{sundaram_dense_2010,
	address = {Berlin, Heidelberg},
	title = {Dense {Point} {Trajectories} by {GPU}-{Accelerated} {Large} {Displacement} {Optical} {Flow}},
	isbn = {978-3-642-15549-9},
	doi = {10.1007/978-3-642-15549-9_32},
	language = {en},
	booktitle = {Computer {Vision} – {ECCV} 2010},
	publisher = {Springer},
	author = {Sundaram, Narayanan and Brox, Thomas and Keutzer, Kurt},
	editor = {Daniilidis, Kostas and Maragos, Petros and Paragios, Nikos},
	year = {2010},
	pages = {438--451},
}

@inproceedings{black_framework_1993,
	title = {A framework for the robust estimation of optical flow},
	url = {https://ieeexplore.ieee.org/abstract/document/378214},
	doi = {10.1109/ICCV.1993.378214},
	urldate = {2024-11-09},
	booktitle = {1993 (4th) {International} {Conference} on {Computer} {Vision}},
	author = {Black, M.J. and Anandan, P.},
	month = may,
	year = {1993},
	pages = {231--236},
}

@misc{fischer_flownet_2015,
	title = {{FlowNet}: {Learning} {Optical} {Flow} with {Convolutional} {Networks}},
	shorttitle = {{FlowNet}},
	url = {http://arxiv.org/abs/1504.06852},
	doi = {10.48550/arXiv.1504.06852},
	urldate = {2024-11-09},
	publisher = {arXiv},
	author = {Fischer, Philipp and Dosovitskiy, Alexey and Ilg, Eddy and Häusser, Philip and Hazırbaş, Caner and Golkov, Vladimir and Smagt, Patrick van der and Cremers, Daniel and Brox, Thomas},
	month = may,
	year = {2015},
	note = {arXiv:1504.06852},
}

@inproceedings{weinzaepfel_deepflow_2013,
	address = {Sydney, Australia},
	title = {{DeepFlow}: {Large} {Displacement} {Optical} {Flow} with {Deep} {Matching}},
	isbn = {978-1-4799-2840-8},
	shorttitle = {{DeepFlow}},
	url = {http://ieeexplore.ieee.org/document/6751282/},
	doi = {10.1109/ICCV.2013.175},
	language = {en},
	urldate = {2024-11-09},
	booktitle = {2013 {IEEE} {International} {Conference} on {Computer} {Vision}},
	publisher = {IEEE},
	author = {Weinzaepfel, Philippe and Revaud, Jerome and Harchaoui, Zaid and Schmid, Cordelia},
	month = dec,
	year = {2013},
	pages = {1385--1392},
}

@misc{khirodkar_sapiens_2024,
	title = {Sapiens: {Foundation} for {Human} {Vision} {Models}},
	url = {https://arxiv.org/abs/2408.12569},
	author = {Khirodkar, Rawal and Bagautdinov, Timur and Martinez, Julieta and Zhaoen, Su and James, Austin and Selednik, Peter and Anderson, Stuart and Saito, Shunsuke},
	year = {2024},
	note = {\_eprint: 2408.12569},
}

@article{jiang_instantavatar_2022-1,
	title = {{InstantAvatar}: {Learning} {Avatars} from {Monocular} {Video} in 60 {Seconds}},
	journal = {arXiv},
	author = {Jiang, Tianjian and Chen, Xu and Song, Jie and Hilliges, Otmar},
	year = {2022},
}

@inproceedings{schonberger_pixelwise_2016,
	title = {Pixelwise {View} {Selection} for {Unstructured} {Multi}-{View} {Stereo}},
	booktitle = {European {Conference} on {Computer} {Vision} ({ECCV})},
	author = {Schönberger, Johannes Lutz and Zheng, Enliang and Pollefeys, Marc and Frahm, Jan-Michael},
	year = {2016},
}

@inproceedings{schonberger_structure--motion_2016,
	title = {Structure-from-{Motion} {Revisited}},
	booktitle = {Conference on {Computer} {Vision} and {Pattern} {Recognition} ({CVPR})},
	author = {Schönberger, Johannes Lutz and Frahm, Jan-Michael},
	year = {2016},
}

@misc{luiten_dynamic_2023,
	title = {Dynamic {3D} {Gaussians}: {Tracking} by {Persistent} {Dynamic} {View} {Synthesis}},
	shorttitle = {Dynamic {3D} {Gaussians}},
	url = {http://arxiv.org/abs/2308.09713},
	language = {en},
	urldate = {2024-05-16},
	publisher = {arXiv},
	author = {Luiten, Jonathon and Kopanas, Georgios and Leibe, Bastian and Ramanan, Deva},
	month = aug,
	year = {2023},
	note = {arXiv:2308.09713 [cs]},
}

@misc{guo_motion-aware_2024,
	title = {Motion-aware {3D} {Gaussian} {Splatting} for {Efficient} {Dynamic} {Scene} {Reconstruction}},
	url = {http://arxiv.org/abs/2403.11447},
	doi = {10.48550/arXiv.2403.11447},
	urldate = {2024-05-16},
	publisher = {arXiv},
	author = {Guo, Zhiyang and Zhou, Wengang and Li, Li and Wang, Min and Li, Houqiang},
	month = mar,
	year = {2024},
	note = {arXiv:2403.11447 [cs]},
}

@misc{jung_deformable_2023,
	title = {Deformable {3D} {Gaussian} {Splatting} for {Animatable} {Human} {Avatars}},
	url = {http://arxiv.org/abs/2312.15059},
	doi = {10.48550/arXiv.2312.15059},
	urldate = {2024-05-16},
	publisher = {arXiv},
	author = {Jung, HyunJun and Brasch, Nikolas and Song, Jifei and Perez-Pellitero, Eduardo and Zhou, Yiren and Li, Zhihao and Navab, Nassir and Busam, Benjamin},
	month = dec,
	year = {2023},
	note = {arXiv:2312.15059 [cs]},
}

@misc{peng_neural_2021,
	title = {Neural {Body}: {Implicit} {Neural} {Representations} with {Structured} {Latent} {Codes} for {Novel} {View} {Synthesis} of {Dynamic} {Humans}},
	shorttitle = {Neural {Body}},
	url = {http://arxiv.org/abs/2012.15838},
	doi = {10.48550/arXiv.2012.15838},
	urldate = {2024-05-16},
	publisher = {arXiv},
	author = {Peng, Sida and Zhang, Yuanqing and Xu, Yinghao and Wang, Qianqian and Shuai, Qing and Bao, Hujun and Zhou, Xiaowei},
	month = mar,
	year = {2021},
	note = {arXiv:2012.15838 [cs]},
}

@misc{peng_animatable_2021,
	title = {Animatable {Neural} {Radiance} {Fields} for {Modeling} {Dynamic} {Human} {Bodies}},
	url = {http://arxiv.org/abs/2105.02872},
	doi = {10.48550/arXiv.2105.02872},
	urldate = {2024-05-16},
	publisher = {arXiv},
	author = {Peng, Sida and Dong, Junting and Wang, Qianqian and Zhang, Shangzhan and Shuai, Qing and Zhou, Xiaowei and Bao, Hujun},
	month = oct,
	year = {2021},
	note = {arXiv:2105.02872 [cs]},
}

@misc{jiang_neuman_2022,
	title = {{NeuMan}: {Neural} {Human} {Radiance} {Field} from a {Single} {Video}},
	shorttitle = {{NeuMan}},
	url = {http://arxiv.org/abs/2203.12575},
	doi = {10.48550/arXiv.2203.12575},
	urldate = {2024-05-16},
	publisher = {arXiv},
	author = {Jiang, Wei and Yi, Kwang Moo and Samei, Golnoosh and Tuzel, Oncel and Ranjan, Anurag},
	month = sep,
	year = {2022},
	note = {arXiv:2203.12575 [cs]},
}

@misc{tretschk_non-rigid_2021,
	title = {Non-{Rigid} {Neural} {Radiance} {Fields}: {Reconstruction} and {Novel} {View} {Synthesis} of a {Dynamic} {Scene} {From} {Monocular} {Video}},
	shorttitle = {Non-{Rigid} {Neural} {Radiance} {Fields}},
	url = {http://arxiv.org/abs/2012.12247},
	doi = {10.48550/arXiv.2012.12247},
	urldate = {2024-05-16},
	publisher = {arXiv},
	author = {Tretschk, Edgar and Tewari, Ayush and Golyanik, Vladislav and Zollhöfer, Michael and Lassner, Christoph and Theobalt, Christian},
	month = aug,
	year = {2021},
	note = {arXiv:2012.12247 [cs]},
}

@misc{cao_hexplane_2023,
	title = {{HexPlane}: {A} {Fast} {Representation} for {Dynamic} {Scenes}},
	shorttitle = {{HexPlane}},
	url = {http://arxiv.org/abs/2301.09632},
	urldate = {2024-05-16},
	publisher = {arXiv},
	author = {Cao, Ang and Johnson, Justin},
	month = mar,
	year = {2023},
	note = {arXiv:2301.09632 [cs]},
}

@misc{wang_fourier_2022,
	title = {Fourier {PlenOctrees} for {Dynamic} {Radiance} {Field} {Rendering} in {Real}-time},
	url = {http://arxiv.org/abs/2202.08614},
	urldate = {2024-05-16},
	publisher = {arXiv},
	author = {Wang, Liao and Zhang, Jiakai and Liu, Xinhang and Zhao, Fuqiang and Zhang, Yanshun and Zhang, Yingliang and Wu, Minye and Xu, Lan and Yu, Jingyi},
	month = feb,
	year = {2022},
	note = {arXiv:2202.08614 [cs]},
}

@misc{huang_one-shot_2023,
	title = {One-shot {Implicit} {Animatable} {Avatars} with {Model}-based {Priors}},
	url = {http://arxiv.org/abs/2212.02469},
	urldate = {2024-05-15},
	publisher = {arXiv},
	author = {Huang, Yangyi and Yi, Hongwei and Liu, Weiyang and Wang, Haofan and Wu, Boxi and Wang, Wenxiao and Lin, Binbin and Zhang, Debing and Cai, Deng},
	month = sep,
	year = {2023},
	note = {arXiv:2212.02469 [cs]},
}

@misc{svitov_haha_2024,
	title = {{HAHA}: {Highly} {Articulated} {Gaussian} {Human} {Avatars} with {Textured} {Mesh} {Prior}},
	shorttitle = {{HAHA}},
	url = {http://arxiv.org/abs/2404.01053},
	urldate = {2024-05-01},
	publisher = {arXiv},
	author = {Svitov, David and Morerio, Pietro and Agapito, Lourdes and Del Bue, Alessio},
	month = apr,
	year = {2024},
	note = {arXiv:2404.01053 [cs]},
}

@misc{liu_gva_2024,
	title = {{GVA}: {Reconstructing} {Vivid} {3D} {Gaussian} {Avatars} from {Monocular} {Videos}},
	shorttitle = {{GVA}},
	url = {http://arxiv.org/abs/2402.16607},
	urldate = {2024-04-22},
	publisher = {arXiv},
	author = {Liu, Xinqi and Wu, Chenming and Liu, Jialun and Liu, Xing and Wu, Jinbo and Zhao, Chen and Feng, Haocheng and Ding, Errui and Wang, Jingdong},
	month = mar,
	year = {2024},
	note = {arXiv:2402.16607 [cs]},
}

@misc{shao_splattingavatar_2024,
	title = {{SplattingAvatar}: {Realistic} {Real}-{Time} {Human} {Avatars} with {Mesh}-{Embedded} {Gaussian} {Splatting}},
	shorttitle = {{SplattingAvatar}},
	url = {http://arxiv.org/abs/2403.05087},
	urldate = {2024-04-22},
	publisher = {arXiv},
	author = {Shao, Zhijing and Wang, Zhaolong and Li, Zhuang and Wang, Duotun and Lin, Xiangru and Zhang, Yu and Fan, Mingming and Wang, Zeyu},
	month = mar,
	year = {2024},
	note = {arXiv:2403.05087 [cs]},
}

@misc{zhao_particlesfm_2022,
	title = {{ParticleSfM}: {Exploiting} {Dense} {Point} {Trajectories} for {Localizing} {Moving} {Cameras} in the {Wild}},
	shorttitle = {{ParticleSfM}},
	url = {http://arxiv.org/abs/2207.09137},
	urldate = {2024-04-18},
	publisher = {arXiv},
	author = {Zhao, Wang and Liu, Shaohui and Guo, Hengkai and Wang, Wenping and Liu, Yong-Jin},
	month = jul,
	year = {2022},
	note = {arXiv:2207.09137 [cs]},
}

@misc{lei_gart_2023,
	title = {{GART}: {Gaussian} {Articulated} {Template} {Models}},
	shorttitle = {{GART}},
	url = {http://arxiv.org/abs/2311.16099},
	urldate = {2024-04-18},
	publisher = {arXiv},
	author = {Lei, Jiahui and Wang, Yufu and Pavlakos, Georgios and Liu, Lingjie and Daniilidis, Kostas},
	month = nov,
	year = {2023},
	note = {arXiv:2311.16099 [cs]},
}

@misc{bae_per-gaussian_2024,
	title = {Per-{Gaussian} {Embedding}-{Based} {Deformation} for {Deformable} {3D} {Gaussian} {Splatting}},
	url = {http://arxiv.org/abs/2404.03613},
	urldate = {2024-04-18},
	publisher = {arXiv},
	author = {Bae, Jeongmin and Kim, Seoha and Yun, Youngsik and Lee, Hahyun and Bang, Gun and Uh, Youngjung},
	month = apr,
	year = {2024},
	note = {arXiv:2404.03613 [cs]},
}

@misc{qian_3dgs-avatar_2023,
	title = {{3DGS}-{Avatar}: {Animatable} {Avatars} via {Deformable} {3D} {Gaussian} {Splatting}},
	shorttitle = {{3DGS}-{Avatar}},
	url = {http://arxiv.org/abs/2312.09228},
	urldate = {2024-03-10},
	publisher = {arXiv},
	author = {Qian, Zhiyin and Wang, Shaofei and Mihajlovic, Marko and Geiger, Andreas and Tang, Siyu},
	month = dec,
	year = {2023},
	note = {arXiv:2312.09228 [cs]},
}

@misc{lin_gaussian-flow_2023,
	title = {Gaussian-{Flow}: {4D} {Reconstruction} with {Dynamic} {3D} {Gaussian} {Particle}},
	shorttitle = {Gaussian-{Flow}},
	url = {http://arxiv.org/abs/2312.03431},
	urldate = {2024-03-10},
	publisher = {arXiv},
	author = {Lin, Youtian and Dai, Zuozhuo and Zhu, Siyu and Yao, Yao},
	month = dec,
	year = {2023},
	note = {arXiv:2312.03431 [cs]},
}

@misc{hu_gaussianavatar_2023,
	title = {{GaussianAvatar}: {Towards} {Realistic} {Human} {Avatar} {Modeling} from a {Single} {Video} via {Animatable} {3D} {Gaussians}},
	shorttitle = {{GaussianAvatar}},
	url = {http://arxiv.org/abs/2312.02134},
	urldate = {2024-03-10},
	publisher = {arXiv},
	author = {Hu, Liangxiao and Zhang, Hongwen and Zhang, Yuxiang and Zhou, Boyao and Liu, Boning and Zhang, Shengping and Nie, Liqiang},
	month = dec,
	year = {2023},
	note = {arXiv:2312.02134 [cs]},
}

@article{loper_smpl_2015,
	title = {{SMPL}: a skinned multi-person linear model},
	volume = {34},
	issn = {0730-0301, 1557-7368},
	shorttitle = {{SMPL}},
	url = {https://dl.acm.org/doi/10.1145/2816795.2818013},
	doi = {10.1145/2816795.2818013},
	language = {en},
	number = {6},
	urldate = {2024-03-10},
	journal = {ACM Transactions on Graphics},
	author = {Loper, Matthew and Mahmood, Naureen and Romero, Javier and Pons-Moll, Gerard and Black, Michael J.},
	month = nov,
	year = {2015},
	pages = {1--16},
}

@misc{yang_real-time_2024,
	title = {Real-time {Photorealistic} {Dynamic} {Scene} {Representation} and {Rendering} with {4D} {Gaussian} {Splatting}},
	url = {http://arxiv.org/abs/2310.10642},
	urldate = {2024-03-10},
	publisher = {arXiv},
	author = {Yang, Zeyu and Yang, Hongye and Pan, Zijie and Zhang, Li},
	month = feb,
	year = {2024},
	note = {arXiv:2310.10642 [cs]},
}

@misc{li_animatable_2023,
	title = {Animatable {Gaussians}: {Learning} {Pose}-dependent {Gaussian} {Maps} for {High}-fidelity {Human} {Avatar} {Modeling}},
	shorttitle = {Animatable {Gaussians}},
	url = {http://arxiv.org/abs/2311.16096},
	urldate = {2024-03-10},
	publisher = {arXiv},
	author = {Li, Zhe and Zheng, Zerong and Wang, Lizhen and Liu, Yebin},
	month = nov,
	year = {2023},
	note = {arXiv:2311.16096 [cs]},
}

@misc{yang_deformable_2023,
	title = {Deformable {3D} {Gaussians} for {High}-{Fidelity} {Monocular} {Dynamic} {Scene} {Reconstruction}},
	url = {http://arxiv.org/abs/2309.13101},
	urldate = {2024-03-10},
	publisher = {arXiv},
	author = {Yang, Ziyi and Gao, Xinyu and Zhou, Wen and Jiao, Shaohui and Zhang, Yuqing and Jin, Xiaogang},
	month = nov,
	year = {2023},
	note = {arXiv:2309.13101 [cs]},
}

@misc{li_spacetime_2023,
	title = {Spacetime {Gaussian} {Feature} {Splatting} for {Real}-{Time} {Dynamic} {View} {Synthesis}},
	url = {http://arxiv.org/abs/2312.16812},
	urldate = {2024-03-10},
	publisher = {arXiv},
	author = {Li, Zhan and Chen, Zhang and Li, Zhong and Xu, Yi},
	month = dec,
	year = {2023},
	note = {arXiv:2312.16812 [cs]},
}

@misc{kocabas_hugs_2023,
	title = {{HUGS}: {Human} {Gaussian} {Splats}},
	shorttitle = {{HUGS}},
	url = {http://arxiv.org/abs/2311.17910},
	urldate = {2024-03-10},
	publisher = {arXiv},
	author = {Kocabas, Muhammed and Chang, Jen-Hao Rick and Gabriel, James and Tuzel, Oncel and Ranjan, Anurag},
	month = nov,
	year = {2023},
	note = {arXiv:2311.17910 [cs]},
}

@misc{wu_4d_2023,
	title = {{4D} {Gaussian} {Splatting} for {Real}-{Time} {Dynamic} {Scene} {Rendering}},
	url = {http://arxiv.org/abs/2310.08528},
	urldate = {2024-02-25},
	publisher = {arXiv},
	author = {Wu, Guanjun and Yi, Taoran and Fang, Jiemin and Xie, Lingxi and Zhang, Xiaopeng and Wei, Wei and Liu, Wenyu and Tian, Qi and Wang, Xinggang},
	month = dec,
	year = {2023},
	note = {arXiv:2310.08528 [cs]},
}

@misc{li_gaussianbody_2024,
	title = {{GaussianBody}: {Clothed} {Human} {Reconstruction} via 3d {Gaussian} {Splatting}},
	shorttitle = {{GaussianBody}},
	url = {http://arxiv.org/abs/2401.09720},
	urldate = {2024-02-15},
	publisher = {arXiv},
	author = {Li, Mengtian and Yao, Shengxiang and Xie, Zhifeng and Chen, Keyu},
	month = jan,
	year = {2024},
	note = {arXiv:2401.09720 [cs]},
}

@misc{kerbl_3d_2023,
	title = {{3D} {Gaussian} {Splatting} for {Real}-{Time} {Radiance} {Field} {Rendering}},
	url = {http://arxiv.org/abs/2308.04079},
	urldate = {2023-12-08},
	publisher = {arXiv},
	author = {Kerbl, Bernhard and Kopanas, Georgios and Leimkühler, Thomas and Drettakis, George},
	month = aug,
	year = {2023},
	note = {arXiv:2308.04079 [cs]},
}

@inproceedings{paszke_pytorch_2019,
	title = {{PyTorch}: {An} {Imperative} {Style}, {High}-{Performance} {Deep} {Learning} {Library}},
	volume = {32},
	shorttitle = {{PyTorch}},
	url = {https://proceedings.neurips.cc/paper/2019/hash/bdbca288fee7f92f2bfa9f7012727740-Abstract.html},
	urldate = {2023-06-09},
	booktitle = {Advances in {Neural} {Information} {Processing} {Systems}},
	publisher = {Curran Associates, Inc.},
	author = {Paszke, Adam and Gross, Sam and Massa, Francisco and Lerer, Adam and Bradbury, James and Chanan, Gregory and Killeen, Trevor and Lin, Zeming and Gimelshein, Natalia and Antiga, Luca and Desmaison, Alban and Kopf, Andreas and Yang, Edward and DeVito, Zachary and Raison, Martin and Tejani, Alykhan and Chilamkurthy, Sasank and Steiner, Benoit and Fang, Lu and Bai, Junjie and Chintala, Soumith},
	year = {2019},
}

@misc{liu2024modgsdynamicgaussiansplatting,
      title={MoDGS: Dynamic Gaussian Splatting from Casually-captured Monocular Videos}, 
      author={Qingming Liu and Yuan Liu and Jiepeng Wang and Xianqiang Lyv and Peng Wang and Wenping Wang and Junhui Hou},
      year={2024},
      eprint={2406.00434},
      archivePrefix={arXiv},
      primaryClass={cs.CV},
      url={https://arxiv.org/abs/2406.00434}, 
}

@misc{lei2024moscadynamicgaussianfusion,
      title={MoSca: Dynamic Gaussian Fusion from Casual Videos via 4D Motion Scaffolds}, 
      author={Jiahui Lei and Yijia Weng and Adam Harley and Leonidas Guibas and Kostas Daniilidis},
      year={2024},
      eprint={2405.17421},
      archivePrefix={arXiv},
      primaryClass={cs.CV},
      url={https://arxiv.org/abs/2405.17421}, 
}

@misc{jiang2021learningopticalflowmatches,
      title={Learning Optical Flow from a Few Matches}, 
      author={Shihao Jiang and Yao Lu and Hongdong Li and Richard Hartley},
      year={2021},
      eprint={2104.02166},
      archivePrefix={arXiv},
      primaryClass={cs.CV},
      url={https://arxiv.org/abs/2104.02166}, 
}

@misc{jiang2021learningestimatehiddenmotions,
      title={Learning to Estimate Hidden Motions with Global Motion Aggregation}, 
      author={Shihao Jiang and Dylan Campbell and Yao Lu and Hongdong Li and Richard Hartley},
      year={2021},
      eprint={2104.02409},
      archivePrefix={arXiv},
      primaryClass={cs.CV},
      url={https://arxiv.org/abs/2104.02409}, 
}

@article{ye2024gsplatopensourcelibrarygaussian,
    title={gsplat: An Open-Source Library for {Gaussian} Splatting}, 
    author={Vickie Ye and Ruilong Li and Justin Kerr and Matias Turkulainen and Brent Yi and Zhuoyang Pan and Otto Seiskari and Jianbo Ye and Jeffrey Hu and Matthew Tancik and Angjoo Kanazawa},
    year={2024},
    eprint={2409.06765},
    journal={arXiv preprint arXiv:2409.06765},
    archivePrefix={arXiv},
    primaryClass={cs.CV},
    url={https://arxiv.org/abs/2409.06765}, 
}

@InProceedings{KingBa15,
  author    = {Kingma, Diederik and Ba, Jimmy},
  booktitle = {International Conference on Learning Representations (ICLR)},
  title     = {Adam: A Method for Stochastic Optimization},
  year      = {2015},
  address   = {San Diega, CA, USA},
  optmonth  = {12},
}

@misc{godard2017unsupervisedmonoculardepthestimation,
      title={Unsupervised Monocular Depth Estimation with Left-Right Consistency}, 
      author={Clément Godard and Oisin Mac Aodha and Gabriel J. Brostow},
      year={2017},
      eprint={1609.03677},
      archivePrefix={arXiv},
      primaryClass={cs.CV},
      url={https://arxiv.org/abs/1609.03677}, 
}

@misc{doersch2024bootstapbootstrappedtrainingtrackinganypoint,
      title={BootsTAP: Bootstrapped Training for Tracking-Any-Point}, 
      author={Carl Doersch and Pauline Luc and Yi Yang and Dilara Gokay and Skanda Koppula and Ankush Gupta and Joseph Heyward and Ignacio Rocco and Ross Goroshin and João Carreira and Andrew Zisserman},
      year={2024},
      eprint={2402.00847},
      archivePrefix={arXiv},
      primaryClass={cs.CV},
      url={https://arxiv.org/abs/2402.00847}, 
}
